\documentclass[twoside]{article}

\usepackage{microtype}
\usepackage{graphicx}
\usepackage{subfigure}
\usepackage{booktabs} 
\usepackage{hyperref}
\usepackage{url}
\usepackage{graphicx}
\usepackage{hyperref}
\usepackage{bm}
\usepackage{enumitem}
\usepackage{graphicx}
\usepackage{amsmath}
\usepackage{subfigure}
\usepackage{array}
\usepackage{booktabs} 
\usepackage{multirow}
\usepackage{caption}
\usepackage{amssymb}
\usepackage{amsmath}
\usepackage{amsthm}
\usepackage{graphicx,subfigure}
\usepackage{hyperref}
\usepackage{xcolor}
\usepackage{xr}

\usepackage[accepted]{aistats2021}

\usepackage[round]{natbib}

\newtheorem{theorem}{Theorem}

\newtheorem{definition}{Definition}
\newtheorem{remark}{Remark}

\newtheorem{corollary}{Corollary}[theorem]

\newcommand*{\colorboxed}{}
\def\colorboxed#1#{%
  \colorboxedAux{#1}%
}
\newcommand*{\colorboxedAux}[3]{%
  \begingroup
    \colorlet{cb@saved}{.}%
    \color#1{#2}%
    \boxed{%
      \color{cb@saved}%
      #3%
    }%
  \endgroup
}
\def\mathunderline#1#2{\color{#1}\underline{{\color{black}#2}}\color{black}}
\definecolor{color1}{HTML}{2E86C1}
\begin{document}

%
\runningauthor{S. Xie*, S. Hu*,  X. Wang, C. Liu, J. Shi, X. Liu, D. Lin}

\twocolumn[

\aistatstitle{Understanding the wiring evolution in differentiable neural architecture search}

\aistatsauthor{Sirui Xie*\And Shoukang Hu*}
\aistatsaddress{University of California, Los Angeles\\ srxie@ucla.edu\And The Chinese University of Hong Kong\\ skhu@se.cuhk.edu.hk}

\aistatsauthor{Xinjiang Wang\And Chunxiao Liu \And Jianping Shi}
\aistatsaddress{SenseTime Research\And SenseTime Research\And SenseTime Research}

\aistatsauthor{Xunying Liu\And Dahua Lin}
\aistatsaddress{The Chinese University of Hong Kong\And The Chinese University of Hong Kong}

]

\begin{abstract}
Controversy exists on whether differentiable neural architecture search methods discover wiring topology effectively. To understand how wiring topology evolves, we study the underlying mechanism of several leading differentiable NAS frameworks. Our investigation is motivated by three observed searching patterns of differentiable NAS: 1) they search by growing instead of pruning; 2) wider networks are more preferred than deeper ones; 3) no edges are selected in bi-level optimization. To anatomize these phenomena, we propose a unified view on searching algorithms of existing frameworks, transferring the global optimization to local cost minimization. Based on this reformulation, we conduct empirical and theoretical analyses, revealing implicit biases in the cost's assignment mechanism and evolution dynamics that cause the observed phenomena. These biases indicate strong discrimination towards certain topologies. To this end, we pose questions that future differentiable methods for neural wiring discovery need to confront, hoping to evoke a discussion and rethinking on how much bias has been enforced implicitly in existing NAS methods. 
\end{abstract}

\section{Introduction}
Neural architecture search (NAS) aims to design neural architecture in an automatic manner, eliminating efforts in heuristics-based architecture design. Its paradigm in the past was mainly black-box optimization \citep{stanley2002evolving, zoph2016neural, kandasamy2018neural}. Failing to utilize the derivative information in the objective makes them computationally demanding. Recently, differentiable architecture search \citep{liu2018darts, xie2018snas, cai2018proxylessnas} came in as a much more efficient alternative, obtaining popularity once at their proposal. These methods can be seamlessly plugged into the computational graph for any generic loss, for which the convenience of automated differentiation infrastructure can be leveraged. They have also exhibited impressive versatility in searching not only operation types but also kernel size, channel number \citep{peng2019efficient}, even wiring topology \citep{liu2018darts, xie2018snas}. 

\begin{figure}[t]
    \centering
    \includegraphics[width=3.0in]{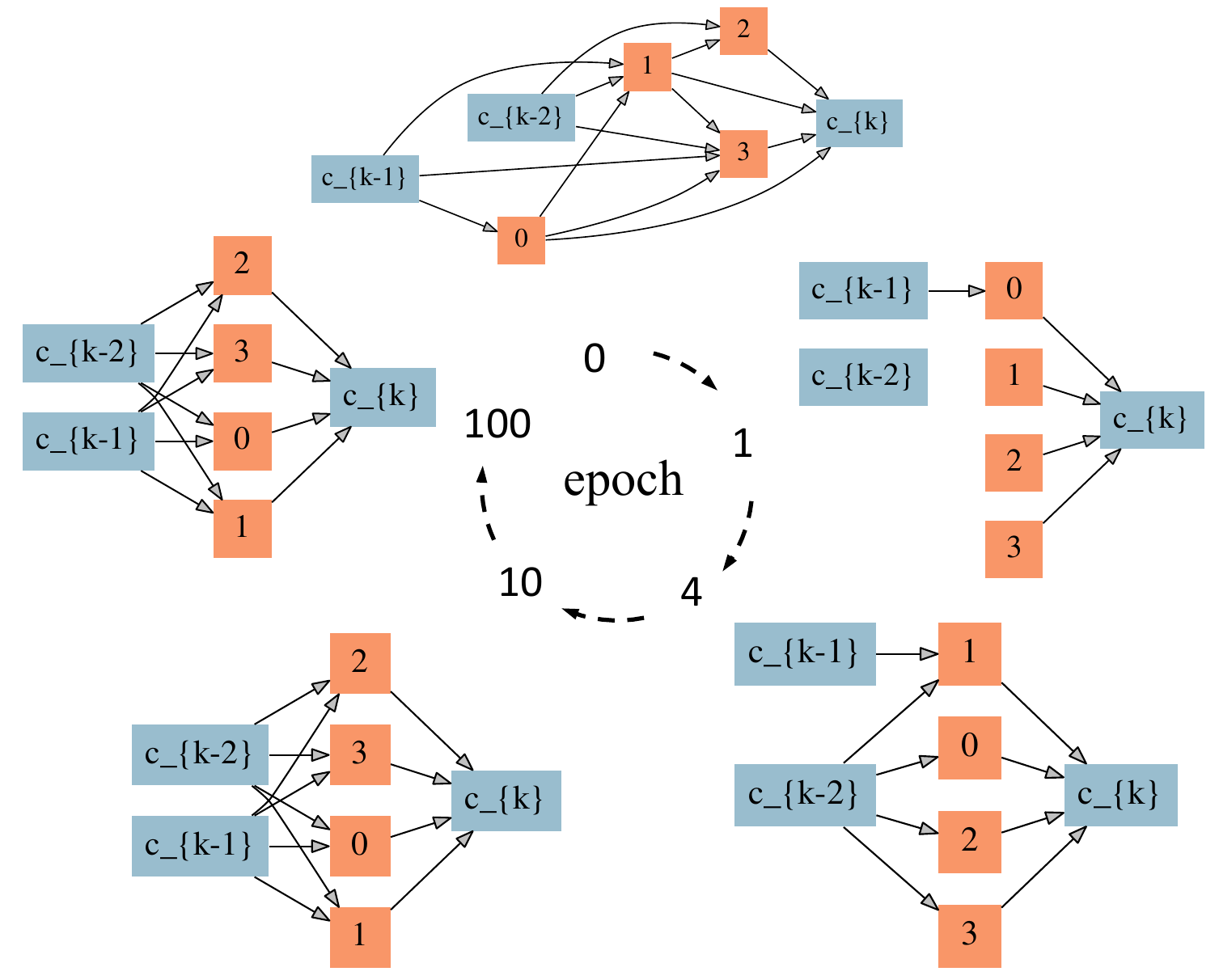}
    \caption{Evolution of the wiring topology with the largest probability in SNAS. Edges are \textit{dropped} in the beginning. Final cells show preference towards width.}
    \label{fig:evolve}
\end{figure}

The effectiveness of wiring discovery in differentiable architecture search, however, is controversial. Opponents argue that the topology discovered is strongly constrained by the manual design of the supernetwork \citep{xie2019exploring}. And it becomes more speculative if ones take a close look at the evolution process, as shown in Fig.\ref{fig:evolve}. From the view of networks with the largest probability, all edges tend to be \textit{dropped} in the beginning, after which some edges recover. The recovery seems slower in depth-encouraging edges and final cells show preference towards width. These patterns occur under variation of search space and search algorithms, to be elaborated in Sec.\ref{sec:phenomena}. A natural question is, if there exists such a universal regularity, why bother searching the topology? 

In this work, we study the underlying mechanism of this wiring evolution. We start from unifying existing differentiable NAS frameworks, transferring NAS to a local cost minimization problem at each edge. As the decision variable at each edge is enumerable, this reformulation makes it possible for an analytical endeavor. Based on it, we introduce a scalable Monte Carlo estimator of the cost that empowers us with a microscopic view of the evolution. Inspired by observations at this level, we conduct empirical and theoretical study on the cost assignment mechanism and the cost evolution dynamics, revealing their relation with the surrogate loss. Based on the results and analysis, we conclude that patterns above are caused (at least partially) by inductive biases in the training/searching scheme of existing NAS frameworks. In other words, there exists implicit discrimination towards certain types of wiring topology that are not necessarily worse than others if trained alone. This discovery poses challenges to future proposals of differentiable neural wiring discovery. To this end, we want to evoke a discussion and rethinking on how much bias has been enforced implicitly in existing neural architecture search methods.\footnote{We have released our code at \url{https://github.com/SNAS-Series/SNAS-Series/tree/master/Analysis}.}

\section{Background}

\subsection{Search Space as Directed Acyclic Graph}
\label{sec:dag}

\begin{figure}[t]
    \centering
    \includegraphics[width=2.6in]{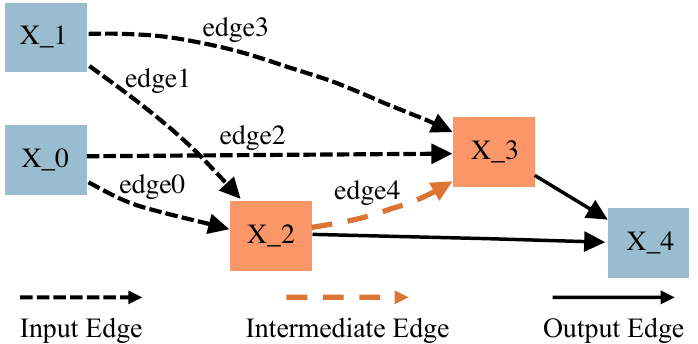}
    \caption{A minimal cell is a cell with 2 \textit{intermediate nodes} (orange), two \textit{input nodes} and one \textit{output node} (blue). Edges connecting input nodes and intermediate nodes ($edge 0-edge3$) are called \textit{input edges}; edges between two intermediate nodes are \textit{intermediate edges} ($edge 4$); others are \textit{output edges} which are skip-connections concatenated together.}
    \label{fig:minimal_cell}
\end{figure}

The recent trend to automatically search for neural architectures starts from \citet{zoph2016neural}. They proposed to model the process of constructing a neural network as a Markov Decision Process, whose policy is optimized towards architectures with high accuracy. Then they moved their focus to the design of search space and proposed ENAS \citep{pham2018efficient}. Particularly, they introduced a Directed Acyclic Graph (DAG) representation of the search space, which is followed by later cell-based NAS frameworks \citep{liu2018darts, xie2018snas}. Since this DAG is one of the main controversial points on the efficacy of differentiable neural wiring discovery, we provide a comprehensive introduction to fill readers in. Fig.\ref{fig:minimal_cell} illustrates a \textit{minimal cell} containing all distinct components. 

Nodes $X_{i}$ in this DAG represent feature maps. Two different types of nodes are colored differently in Fig.\ref{fig:minimal_cell}. \textit{Input nodes} and \textit{output nodes} (blue) are nodes indicating the inter-connection between cells. Here for instance, $X_0$ represents both the input to this cell and the output from the previous cell. Similarly, $X_1$ represents both the input to this cell and the output from the one before the previous cell. In symmetry, $X_4$ is the output of this cell, and is also the input to some next cells. Between input nodes and output nodes, there are \textit{intermediate nodes} (orange). Numbers of each type of nodes are hyper-parameters of this design scheme. 

Edges $(i, j)$ represent information flows between nodes $X_{i}$ and $X_{j}$. Every intermediate node is connected to all previous nodes with smaller indices. If the edge is between an input node and an intermediate node, it is called \textit{input edge}. If its vertices are both intermediate nodes, it is called \textit{intermediate edge}. The rest are \textit{output} edges, concatenating all intermediate nodes together to the output node. Notice that all edges except output edges are compounded with a list of operation candidates $\bm{O}_{i, j}$. Normally, candidates include \textit{convolution}, \textit{pooling} and \textit{skip-connection}, augmented with a \textit{ReLU-Conv-BN} or \textit{Pool-BN} stacking. 

\subsection{Differentiable Neural Architecture Search}
\label{sec:nas}

Model-free policy search methods in ENAS leave over the derivative information in network performance evaluators and knowledge about the transition model, two important factors that can boost the optimization efficiency. Building upon this insight, differentiable NAS frameworks are proposed \citep{liu2018darts, xie2018snas}. Their key contributions are novel differentiable instantiations of the DAG, representing not only the search space but also the network construction process. Notice that the operation selection process can be materialized as 
\begin{equation}
\tilde{\bm{O}}_{i, j}(\cdot)= \bm{Z}_{i,j}^{T}\bm{O}_{i, j}(\cdot),
\label{eq:compound_edge}
\end{equation}
\textit{i.e.} multiplying a one-hot random variable $\bm{Z}_{i, j}$ to the output $\bm{O}_{i, j}(\cdot)$ of each edge $(i, j)$. And if we further replace the non-differentiable network accuracy with differentiable surrogate loss, \textit{e.g.} training loss or testing loss, the NAS objective would become 
\begin{equation}
 \mathbb{E}_{\bm{Z}\sim p_{\bm{\alpha}}(\bm{Z})}[L_{\bm{{\bm{\theta}}}}(\bm{Z})], 
\label{eq:objective}
\end{equation}
where $\bm{\alpha}$ and ${\bm{\theta}}$ are parameters of architecture distribution and operations respectively, and $L_{\bm{{\bm{\theta}}}}$ is the surrogate loss, which can be calculated on either training sets as in single-level optimization or testing sets in bi-level optimization. For classification,
\begin{equation}
L_{\bm{{\bm{\theta}}}} = \mathbb{E}_{(x^{b}, y^{b}) \sim D}[CE(g_{\bm{{\bm{\theta}}}}(x^{b}), y^{b})],
\label{eq:cross_ent}
\end{equation}
where $D$ denotes dataset, $(x^{b}, y^{b})$ is $b$-th pair of input and label, $g_{\bm{{\bm{\theta}}}}$ is the output of the sampled network, $CE$ is the cross entropy loss. 

DARTS \citep{liu2018darts} and SNAS \citep{xie2018snas} deviate from each other on how to make $p_{\bm{\alpha}}$ in $\mathbb{E}_{\bm{Z}\sim p_{\bm{\alpha}}}[.]$ differentiable. DARTS relaxes Eq.\ref{eq:compound_edge} continuously with deterministic weights $\hat{\bm{Z}}_{i, j}^{k}$. SNAS keeps the network sampling process represented by Eq.\ref{eq:compound_edge}, while making it differentiable with the softened one-hot random variable $\tilde{\bm{Z}}_{i, j}$ from the gumbel-softmax technique:
\begin{equation*}
\begin{split}
\label{eq:softmax}
\hat{\bm{Z}}_{i, j}^{k} &= \frac{e^{(\log \bm{\alpha}_{i, j}^{k})/\lambda}}{\sum_{l=0}^{n}e^{ (\log \bm{\alpha}_{i, j}^{l})/\lambda}}, \tilde{\bm{Z}}_{i, j}^{k} = \frac{e^{(\log \bm{\alpha}_{i, j}^{k} + \bm{G}_{i, j}^{k})/\lambda}}{\sum_{l=0}^{n}e^{(\log \bm{\alpha}_{i, j}^{l} + \bm{G}_{i, j}^{l})/\lambda}},
\end{split}
\end{equation*}
where $\lambda$ is the temperature of the softmax, $\bm{G}_{i, j}^{k} = -\log (-\log (\bm{U}_{i, j}^{k}))$ is the $k$-th \textit{Gumbel} random variable, $\bm{U}_{i, j}^{k}$ is a uniform random variable. A detailed discussion on SNAS's benefits over DARTS from this sampling process is provided in \citet{xie2018snas}.

\subsection{Neural Wiring Discovery}
\label{sec:wiring}

The sophisticated design scheme in Sec.\ref{sec:dag} is expected to serve the purpose of neural wiring discovery, more specifically, simultaneous search of operation and topology, two crucial aspects in neural architecture design \citep{he2016deep}. Back to the example in Fig.\ref{fig:minimal_cell}, if the optimized architecture does not pick any candidate at all \textit{input edges} from certain \textit{input nodes}, \textit{e.g.} $edge1$ and $edge3$, it can be regarded as a simplified inter-cell wiring structure is discovered. Similarly, if the optimized architecture does not pick any candidate at certain \textit{intermediate edges}, a simplified intra-cell wiring structure is discovered. In differentiable NAS frameworks \citep{liu2018darts, xie2018snas}, this is achieved by adding a \textit{None} operation to the candidate list. The hope is that finally some \textit{None} operations will be selected.  

However, effectiveness of differentiable neural wiring discovery has become speculative recently. On the one hand, only searching for operations in the single-path setting such as ProxylessNAS \citep{cai2018proxylessnas} and SPOS \citep{guo2019single} reaches state-of-the-art performance, without significant modification on the overall training scheme. \citet{xie2019exploring} searches solely for topology with random graph generators, achieves even better performance. On the other hand, \textit{post hoc} analysis has shown that networks with width-preferred topology similar to those discovered in DARTS and SNAS are easier to optimize comparing with their depth-preferred counterparts \citep{shu2019understanding}. Prior to these works, there has been a discussion on the same topic, although from a low-level view. \citet{xie2018snas} points out that the wiring discovery in DARTS is not a direct derivation from the optimized operation weight. Rather, \textit{None} operation is excluded during final operation selection and an \textit{post hoc} scheme to select edges with top-$k$ weight is designed for the wiring structure. They further illustrate the bias of this derivation scheme. Although this bias exists, some recent works have reported impressive results based on DARTS \citep{xu2019pc, chu2019fair}, especially in wiring discovery \citep{wortsman2019discovering}. 

In this work, we conduct a genetic study on differentiable neural wiring discovery. 

\begin{figure}[t]
    \centering
    \includegraphics[width=3.1in]{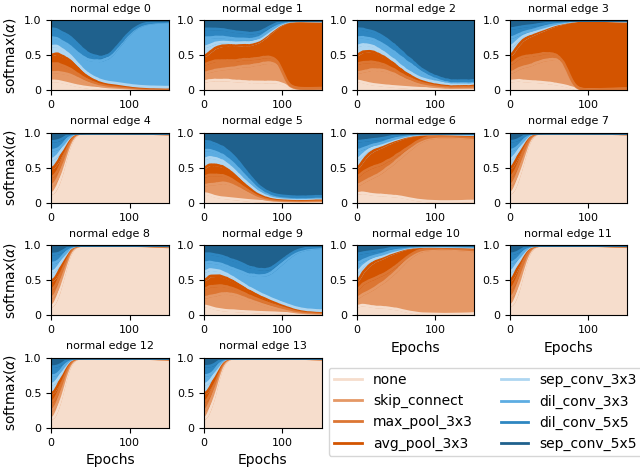}
    \caption{Evolution of $p_{\alpha}$ with epochs on each edge in SNAS (single-level optimization). Different colors are for probabilities of different operations. 7, 8, 11, 12, 13 are intermediate edges where \textit{None} dominates others. This domination is manifested as the \textit{width preference}. The rest are input edges, where \textit{None} only dominates in the beginning. Other operations then become the most likely ones, exhibiting the \textit{growth tendency}.}
    \label{fig:snas_update_alpha}
\end{figure}
 
\section{Motivating observations}
\label{sec:phenomena}

Three evolution patterns occur in networks with largest probability when we replicate experiments of DARTS, SNAS, and DSNAS\footnote{We extend single-path DSNAS to cells. All setting same as in \citet{xie2018snas}, except no resource constraint.} (Fig. \ref{fig:snas_update_alpha}):
\begin{itemize}[leftmargin=*,noitemsep,nolistsep]
    \item (\textbf{P1}) \textit{Growing tendency}: Though starting from a fully-connected super-network, the wiring evolution with single-level optimization is not a pruning process. Rather, all edges are dropped\footnote{\textit{Drop} is soft as it describes the probability of operations.} in the beginning. Some of them then gradually recover.
    \item (\textbf{P2}) \textit{Width preference}: Intermediate edges barely recover from initial drops, even in single-level optimization, unless search spaces are extremely small.
    \item (\textbf{P3}) \textit{Catastrophic failure}: With bi-level optimization, where architecture are updated with held-out data, no edge can recover from the initial drop.
\end{itemize}

These patterns are universal in the sense that they robustly hold with variation of random seeds, learning rate, the number of cells, the number of nodes in a cell, and the number of operation candidates at each edge. More evolution details are provided in Appx.\ref{app:evolution}.

\section{A unified framework for differentiable NAS}

Universal phenomena reported above motivate us to seek for a unified framework. In Sec. \ref{sec:nas}, we unify the formulation of DARTS and SNAS. However, this high-level view does not provide us with much insight on the underlying mechanism of architecture evolution. Apart from this high-level unification, \citet{xie2018snas} also provides a microscopic view by deriving the policy gradient equivalent of SNAS's search gradient
\begin{equation*}
\begin{split}
&\mathbb{E}_{p(\tilde{\bm{Z}})}[\frac{\partial \mathcal{L}}{\partial \alpha_{i,j}^{k}}]=\mathbb{E}_{p(\tilde{\bm{Z}})}[\nabla_{\alpha_{i,j}^{k}}\log p(\tilde{\bm{Z}})[\frac{\partial \mathcal{L}}{\partial x_{j}}\bm{O}_{i,j}^{T}(x_{i}) \tilde{\bm{Z}}_{i, j}]_{c}],
\label{eq:cost_obj_cont_snas}
\end{split}
\end{equation*}
where $\tilde{\bm{Z}}_{i,j}$ is the gumbel-softmax random variable, $[f]_{c}$ denotes that $f$ is a \textit{cost}  independent from $\bm{\alpha}$ for gradient calculation, $\mathcal{L}$ is differentiated from $L$ to highlight the approximation introduced by $\tilde{\bm{Z}}_{i,j}$. Zooming from the global-level objective to a local-level \textit{cost minimization}, it naturally divides the intractable NAS problem into tractable ones, because we only have a manageable set of operation candidates. In this section we generalize it to unify other differentiable NAS frameworks. 

Recently, \citet{hu2020dsnas} proposed to approximate at the discrete limit $\lim_{\lambda \to 0}\tilde{\bm{Z}}={\bm{Z}}$, and the gradient above comes
\begin{equation*}
\begin{split}
\mathbb{E}_{p(\bm{Z})}[\nabla_{\alpha_{i,j}^{k}}\log p(\bm{Z}_{i,j})[\frac{\partial L}{\partial x_{j}}\sum_{k}{O}^{k}_{i, j}(x_{i}){Z}^{k}_{i,j}]_{c}],
\label{eq:snas_disc}
\end{split}
\end{equation*}
where $\bm{Z}_{i,j}$ is a strictly one-hot random variable, ${Z}^{k}_{i,j}$ is the $k$th element in it. Exploiting the one-hot nature of $\bm{Z}_{i,j}$, \textit{i.e.} only $Z^{s}_{i,j}$ on edge $(i, j)$ is 1, others \textit{i.e.} $Z^{\smallsetminus s}_{i,j}$ are $0$, they further reduce the \textit{cost} function to
\begin{equation*}
\begin{split}
C(\bm{Z}_{i,j}) &=\sum_{k}\frac{\partial L}{\partial x_{j}}{O}^{k}_{i, j}(x_{i}){Z}^{k}_{i,j}
 = \frac{\partial L}{\partial x_{j}^{i}}x_{j}^{i} = \frac{\partial L}{\partial Z^{s}_{i,j}}. 
\end{split}
\label{eq:cost_disc}
\end{equation*}
We can also derive the cost function of DARTS and ProxylessNAS in a similar form:
\begin{equation*}
\begin{split}
    \frac{\partial \mathcal{L}^{DARTS}}{\partial \bm{\alpha}_{i,j}^{}} = \frac{\partial \hat{\bm{Z}}_{i,j}^{}}{\partial \bm{\alpha}_{i,j}^{}}C(\hat{\bm{Z}}_{i,j}^{}),  C(\hat{\bm{Z}}_{i,j}^{})=\frac{\partial \mathcal{L}}{\partial \hat{\bm{Z}}_{i,j}^{}},\\
    \quad \frac{\partial \mathcal{L}^{Proxyless}}{\partial \bm{\alpha}_{i,j}^{}} = \frac{\partial \hat{\bm{Z}}_{i,j}^{}}{\partial \bm{\alpha}_{i,j}^{}}C({\bm{Z}}_{i,j}^{}), C(\bm{Z}_{i,j}^{})=\frac{\partial \mathcal{L}}{\partial \bm{Z}_{i,j}^{}},
\label{eq:darts_proxy_general}
\end{split}
\end{equation*}
where $\mathcal{L}^{DARTS}$ and $\mathcal{L}^{Proxyless}$ highlight the loss is an approximation of $\mathbb{E}_{\bm{Z}}[L]$. $\hat{\bm{Z}}_{i,j}$ highlights the attention-based estimator, also referred to as the straight-through technique \citep{bengio2013estimating}. ${\bm{Z}}_{i,j}$ is the discrete selection in Eq.\ref{eq:compound_edge}. Details are provided in Appx.\ref{app:cost_darts_prox}.

Seeing the consistency in these gradients, we reformulate architecture optimization in existing differentiable NAS frameworks as a \textit{local cost optimization} with: 
\begin{equation}
    \label{eq:cost_obj}
    \frac{\partial }{\partial \bm{\alpha}_{i,j}^{}}\mathbb{E}_{\bm{Z}_{i,j}\sim p(\bm{Z}_{i,j})}[C(\bm{Z}_{i,j})], \quad
    C(\bm{Z}_{i,j}) = \frac{\partial \mathcal{L}}{\partial x_{j}^{i}}x_{j}^{i} = \frac{\partial \mathcal{L}}{\partial Z^{s}_{i,j}}.
\end{equation}
Previously, \citet{xie2018snas} interpreted this cost as the first-order Taylor decomposition \citep{montavon2017explaining} of the loss with a convoluted notion, \textit{effective layer}. Aiming at a cleaner perspective, we expand the point-wise form of the cost, with $\bm{X}^{i}_{j}$ denoting the feature map from edge $(i, j)$, $\bm{Z}_{\smallsetminus  i,j}$ denoting marginalizing out all other edges except $(i,j)$, $w,h,c,b$ denoting the width, height, channel and batch size:
\begin{equation}
\begin{split}
C(\bm{Z}_{i,j}) &=\mathbb{E}_{\bm{Z}_{\smallsetminus  i,j}}[\sum_{w,h,c,b} \frac{\partial \mathcal{L}}{\partial [X_{j}^{i}]^{whcb}}[X_{j}^{i}]^{whcb}].
\end{split}
\label{eq:cost_loss_ls}
\end{equation}

\section{Analysis}
It does not come straightforward how this cost minimization (Eq.\ref{eq:cost_obj}) achieves wiring topology discovery. In this section, we conduct an empirical and theoretical study on \textit{the assignment and the dynamics of the cost with the training of operation parameters ${\bm{\theta}}$ and how it drives the evolution of wiring topology}. Our analysis focuses on the appearance and disappearance of \textit{None} operation because the wiring evolution is a manifestation of choosing \textit{None} or not at edges. We start by digging into statistical characteristics of the cost behind phenomena in Sec.\ref{sec:phenomena} (Sec.\ref{sec:stat_of_cost}). We then derive the cost assignment mechanism from output edges to input edges and intermediate edges and introduce the dynamics of the total cost, a strong indicator for the \textit{growing tendency} (Sec.\ref{sec:cost_dist}). By exhibiting the unequal assignment of cost (Sec.\ref{sec:cost_of_inter_edge}), we explain the underlying cause of \textit{width preference}. Finally, we discuss the exceptional cost tendency in bi-level optimization, underpinning \textit{catastrophic failure} (Sec.\ref{sec:bi_level_effect}). Essentially, we reveal the aforementioned patterns are rooted in some implicit biases in existing differentiable NAS frameworks.

\subsection{Statistics of cost}
\label{sec:stat_of_cost}

Since the cost of \textit{None} fixes as $0$, the network topology is determined by signs of cost of other operations, which may change with operation parameters ${\bm{\theta}}$ updated. Though we cannot analytically calculate the expected cost of other operations, we can estimate it with a Monte Carlo simulation.  

To simplify the analysis without lost of generality, we consider DSNAS on one minimal cell with only one intermediate edge (Fig.\ref{fig:minimal_cell}). One cell is sufficient since cost at the last cell dominates with at least one-degree-larger magnitude, which can be explained by the vanishing gradient phenomenon. All experiments reported below are on CIFAR-10 dataset. As will be seen, our analysis is independent of datasets. Monte Carlo simulations are conducted in the following way: After initialization\footnote{$\epsilon$ in the batch normalization is set to be 1e-5.}, subnetworks are sampled uniformly with weight parameters $\bm{{\bm{\theta}}}$ and architecture parameters $\bm{\alpha}$ fixed. The cost of each sampled operation at each edge is stored. After 50 epochs of sampling, statistics of the cost are calculated.

\begin{figure}[t]
    \centering
    \includegraphics[width=3.3in]{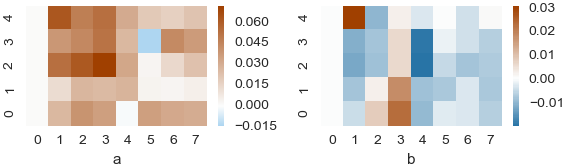}
    \caption{Mean cost for each operation enumerated in x-axis on each edge enumerated in y-axis (a) at initialization and (b) near convergence of $\theta$. Operation 0: none, Operation 1: skip connect, Operation 2: max\_pool\_3x3, Operation 3: avg\_pool\_3x3, Operation 4: sep\_conv\_3x3, Operation 5: dil\_conv\_3x3, Operation 6: dil\_conv\_5x5, Operation 7: sep\_conv\_5x5.}
    \label{fig:cost_mean}
    \vspace{-1em}
\end{figure}
Surprisingly, for all operations except \textit{None}, cost is inclined towards positive at initialization (Fig.\ref{fig:cost_mean}(a)). Similarly, we estimate the mean cost after updating weight parameters $\bm{{\bm{\theta}}}$ for 150 epochs\footnote{Here we strictly follow the training setup in \citep{liu2018darts}, with BN affine~\citep{ioffe2015batch}  disabled.} with architecture parameters $\bm{\alpha}$ still fixed. As shown in Fig.\ref{fig:cost_mean}(b), most of the cost becomes negative. It then becomes apparent that \textit{None} operations are preferred in the beginning as they minimize these costs. After training, the cost minimizer would prefer operations with the smallest negative cost. We formalize this hypothesis as:  

\textbf{Hypothesis}: Cost of operations except \textit{None} are positive \textit{near} initialization. It \textit{eventually} turns negative with the training of operation parameters. Cell topology thus exhibits a tendency of growing.

Though Fig.\ref{fig:cost_mean}(a)(b) show statistics of the minimal cell, this Monte Carlo estimation can scale up to more cells with more nodes. In our analysis, it is a probe for the underlying mechanism. We provide its instantiation in the original training setting \citep{xie2018snas} \textit{i.e.} stacking 8 cells with 4 intermediate nodes in Appx.\ref{app:cost_8_cells}, whose inclination is consistent with the minimal cell here. 

\subsection{The cost assignment mechanism}
\label{sec:cost_dist}

To theoretically validate this hypothesis, we analyze the sign-changing behavior of the cost (Eq.\ref{eq:cost_loss_ls}). We start by listing the circumscription, eliminating possible confusion in derivation below. 
\begin{definition}
Search space $\mathcal{A} = \{$base architecture (DAG) \textbf{A}: \textbf{A} can be constructed as in Sec. \ref{sec:dag}$\}$. 
\end{definition}
\vspace{-0.6em}
Although Eq.\ref{eq:cost_loss_ls} generally applies to search spaces such as cell-based, single-path or modular, in this work we showcase the wiring evolution in $\mathcal{A}$ \textit{w.o.l.g.} We recommend readers to review Sec. \ref{sec:dag}. In particular, we want to remind readers that intermediate edges are those connecting intermediate nodes.


We first study the assignment mechanism of cost, which by intuition is driven by gradient back-propagation. 

There are two types of edges in terms of gradient back-propagation. The more complicated ones are edges pointing to nodes with outgoing intermediate edges, \textit{e.g.}, $edge 0$ and $edge 1$ in Fig. \ref{fig:minimal_cell}. Take $edge 0$ for example, its cost considers two paths of back-propagation, \textit{i.e.}  $1^{st}$ path (4-2-0) and $2^{nd}$ path (4-3-2-0): 
\begin{equation}
\label{eq:cost_edge_0_fir}
\begin{aligned}
    C(Z_{0,2}^{s})&=\frac{\partial L_{\bm{\theta}}}{\partial \bm{X}_{4}^{2}}\frac{\partial \bm{X}_{4}^{2}}{\partial \bm{X}_{2}^{0}} \bm{X}_{2}^{0}+\frac{\partial L_{\bm{\theta}}}{\partial \bm{X}_{4}^{3}}\frac{\partial \bm{X}_{4}^{3}}{\partial \bm{X}_{3}^{2}}\frac{\partial \bm{X}_{3}^{2}}{\partial \bm{X}_{2}^{0}}\bm{X}_{2}^{0} \\
    &=\frac{\partial L_{\bm{\theta}}}{\partial \bm{X}_{4}^{2}} \bm{X}_{2}^{0}+\frac{\partial L_{\bm{\theta}}}{\partial \bm{X}_{3}^{2}}\frac{\partial \bm{X}_{3}^{2}}{\partial \bm{X}_{2}^{0}}\bm{X}_{2}^{0},
\end{aligned}
\end{equation}
where $\frac{\partial L_{\bm{\theta}}}{\partial \bm{X}_{4}^{2}} \bm{X}_{2}^{0}$ and $\frac{\partial L_{\bm{\theta}}}{\partial \bm{X}_{3}^{2}}\frac{\partial \bm{X}_{3}^{2}}{\partial \bm{X}_{2}^{0}}\bm{X}_{2}^{0}$ denote the cost functions calculated from first path and second path respectively, and we have $\bm{X}_{4}^{2}=\bm{X}_{2}^{0}+\bm{X}_{2}^{1}$ at the $1^{st}$ path and $\bm{X}_{4}^{3}=\bm{X}_{3}^{0}+\bm{X}_{3}^{1}+\bm{X}_{3}^{2}$ on the $2^{nd}$ path, $\bm{X}^{i}_{j}$ is result from $edge (i,j)$.

The rest are edges pointing to nodes whose only outgoing edges are output edges, \textit{e.g.} $edge 2$, $edge 3$ and $edge 4$. Take $edge 2$ for example, its cost only involves one path of back-propagation, i.e., (4-3-0): 
\begin{equation}
\begin{split}
\label{eq:cost_edge_0_sec}
    C(Z_{0,3}^{s})&=\frac{\partial L_{\bm{\theta}}}{\partial \bm{X}_{4}^{3}}\frac{\partial \bm{X}_{4}^{3}}{\partial \bm{X}_{3}^{0}} \bm{X}_{3}^{0} =\frac{\partial L_{\bm{\theta}}}{\partial \bm{X}_{4}^{3}} \bm{X}_{3}^{0}.
\end{split}
\end{equation}
Even though Eq.\ref{eq:cost_edge_0_fir} seems to have one more term than Eq.\ref{eq:cost_edge_0_sec} $\frac{\partial L_{\bm{\theta}}}{\partial \bm{X}_{3}^{2}}\frac{\partial \bm{X}_{3}^{2}}{\partial \bm{X}_{2}^{0}}\bm{X}_{2}^{0}$, we can prove that this term equals 0 in the current setup of differentiable NAS: 
\begin{theorem} 
\label{thm:bn0}
A path does not distribute cost from its output edge after passing one intermediate edge.
\end{theorem}
\vspace{-1em}
\begin{proof}(Sketch)
 Let $\bm{X}\in \mathbb{R}^{B\times C_{out} \times W_{out} \times H_{out}}$ denotes the Conv output on edge $4$, we expand $C(Z_{0,2}^{s})$ at path (4-3-2-0): 
\begin{equation*}
\label{eq:cost_edge_01_sec}
\begin{aligned}
    \frac{\partial L_{\bm{\theta}}}{\partial \bm{X}_{3}^{2}}\frac{\partial \bm{X}_{3}^{2}}{\partial \bm{X}_{2}^{0}}\bm{X}_{2}^{0}=\frac{\partial L_{\bm{\theta}}}{\partial \bm{X}_{3}^{2}} \frac{\partial \bm{X}_{3}^{2}}{\partial \bm{X}} \frac{\partial \bm{X}}{\partial \bm{X}_{2}^{0}}\bm{X}_{2}^{0}.
\end{aligned}
\end{equation*}
With element-wise expansion, we can prove  
\begin{equation*}
    \frac{\partial L_{\bm{\theta}}}{\partial \bm{X}_{3}^{2}} \frac{\partial \bm{X}_{3}^{2}}{\partial \bm{X}} \frac{\partial \bm{X}}{\partial \bm{X}_{2}^{0}}\bm{X}_{2}^{0}=\frac{\partial L_{\bm{\theta}}}{\partial \bm{X}_{3}^{2}} \frac{\partial \bm{X}_{3}^{2}}{\partial \bm{X}} \bm{X}.
\end{equation*}
Exploiting the property of normalization, we have
\begin{equation*}
    \frac{\partial L_{\bm{\theta}}}{\partial \bm{X}_{3}^{2}} \frac{\partial \bm{X}_{3}^{2}}{\partial \bm{X}} \bm{X}=0.
\end{equation*}
See Appx.\ref{app:proof_thm2} for the full proof. Note this result can be generalized to arbitrary back-prop path involving intermediate edges. The theorem is thus proved. 
\end{proof}
\vspace{-0.6em}

\begin{figure}[t]
    \centering
    \includegraphics[width=2.6in]{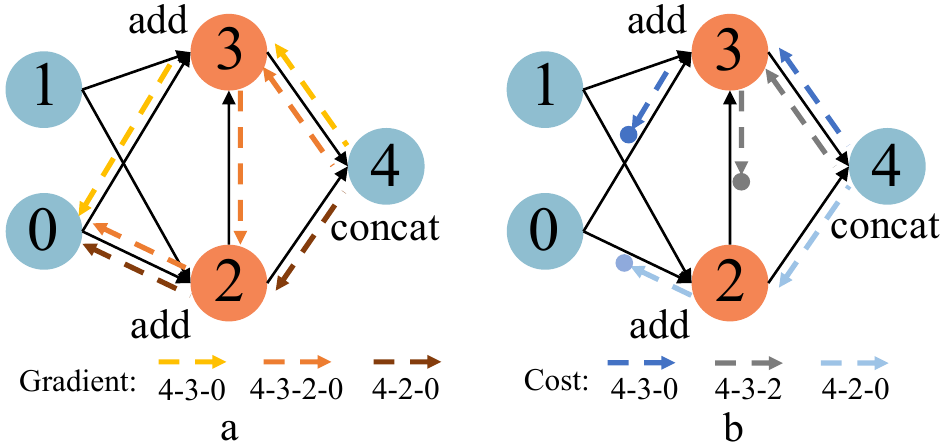}
    \caption{(a) Gradient back-prop in the minimal cell and (b) cost assignment in the minimal cell. Note that their difference on edge $(0,2)$ is caused by the absorption of cost on edge $(2,3)$.}
    \label{fig:cost_gradient}
\vspace{-1em}
\end{figure}
\begin{figure}[t]
    \centering
    \includegraphics[width=2.17in]{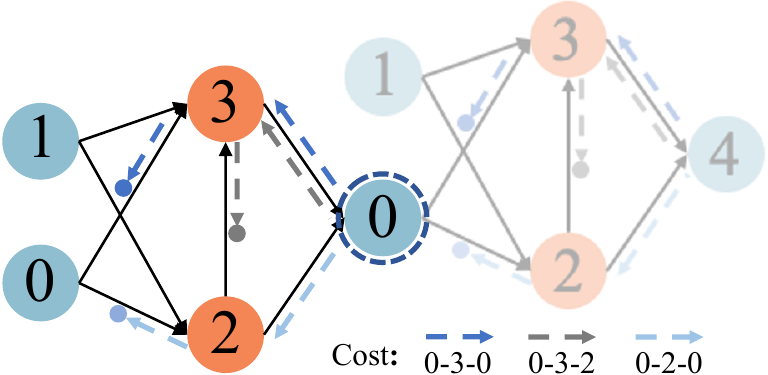}
    \caption{Cost assignment in the last two cells. The cost of the left cell originates from node 0 of the right cell and should sum up to zero. }
    \label{fig:cost_gradient_2layers}
\vspace{-1.5em}
\end{figure}
By Thm.\ref{thm:bn0}, cost on $edge (i,j)$ is only distributed from its subsequent output $edge (j, out)$. Subsequent intermediate $edge (j, k)$ does not contribute to it. As illustrated in Fig.\ref{fig:cost_gradient}(a)(b), Thm.\ref{thm:bn0} reveals the difference between the cost assignment and gradient back-propagation.\footnote{It thus implies that the notion of multiple \textit{effective layers} in \citet{xie2018snas} is an illusion in search space $\mathcal{A}$.} An edge only involves in one layer of cost assignment. With this theorem, the second term $\frac{\partial L_{\bm{\theta}}}{\partial \bm{X}_{3}^{2}}\frac{\partial \bm{X}_{3}^{2}}{\partial \bm{X}_{2}^{0}}\bm{X}_{2}^{0}$ in Eq.\ref{eq:cost_edge_0_fir} can be dropped. Eq.\ref{eq:cost_edge_0_fir} is thus in the same form as Eq.\ref{eq:cost_edge_0_sec}. That is, this theorem brings about a universal form of cost on all edges:

\begin{corollary}
\label{thm:cost_dist}
The total cost distributes to edges in the same cell as 
$$\frac{\partial L_{\bm{\theta}}}{\partial \bm{X}_{output}} \bm{X}_{output}=\sum_j\frac{\partial L_{\bm{\theta}}}{\partial \bm{X}^{j}_{output}} (\sum_{i<j}\bm{X}^{i}_{j}),$$
where $\bm{X}_{output}$ is the output node, $j\in \{$intermediate nodes$\}$, $i\in \{$all nodes$: i<j\}$. 
\end{corollary}
    For example, in the minimal cell above, we have $$\frac{\partial L_{\bm{\theta}}}{\partial \bm{X}_{4}} \bm{X}_{4}=\frac{\partial L_{\bm{\theta}}}{\partial \bm{X}^{3}_{4}} (\bm{X}^{0}_{3}+\bm{X}^{1}_{3}+\bm{X}^{2}_{3})+\frac{\partial L_{\bm{\theta}}}{\partial \bm{X}^{2}_{4}} (\bm{X}^{0}_{2}+\bm{X}^{1}_{2}).$$ In the remaining parts of this paper, we would refer to $\frac{\partial L_{\bm{\theta}}}{\partial \bm{X}_{output}} \bm{X}_{output}$ with \textit{the cost at output nodes}, and refer to $\frac{\partial L_{\bm{\theta}}}{\partial \bm{X}^{j}_{output}} (\sum_{i<j}\bm{X}^{i}_{j})$ with \textit{the cost at intermediate nodes}. Basically these are cost sums of edges pointing to nodes. Even though all cells have this form of cost assignment, we can prove that the total cost in each cell except the last one sum up to 0. More formally:
\begin{corollary}
\label{thm:sum0}
In cells except the last one, for intermediate nodes that are connected to the same output node, cost of edges pointing to them sums up to 0.
\end{corollary}
\vspace{-1em}
\begin{proof}
For all search spaces in $\mathcal{A}$, input nodes of the last cell \textit{e.g} $X_0$, are the output node of previous cells, if they exist (Sec.\ref{sec:dag}). Consider the output node of the second last cell. The cost sum at this node is distributed from the last cell along paths with at least one intermediate node, \textit{e.g.} (4-2-0) and (4-3-2-0), as illustrated in Fig.\ref{fig:cost_gradient_2layers}. By Thm.\ref{thm:bn0}, this cost sum is 0. The same claim for output nodes in other cells can be proved by induction. See Appx.\ref{app:val_cor} for a validation. 
\end{proof}


Let's take a close look at the total cost of the last cell. 
\begin{theorem}
\label{thm:cost_dynamics}
Cost at output edges of the last cell has the form $C_{\bm{Z}}=L_{\bm{\theta}}-H_{\bm{\theta}}$. 
It is negatively related to classification accuracy. It tends to be positive at low accuracy, negative at high accuracy.
\end{theorem}

\begin{proof}(Sketch) \textbf{Negatively related}:
We first prove that for the last cell's output edges, cost of one batch $M$ with sampled architecture $\bm{Z}$ has an equivalent form:
\begin{equation}
\label{eq:cost_ce_ent}
\begin{aligned}
    C_{\bm{Z}}&=\sum_{b,c,d}\frac{\partial{L_{\bm{\theta}}}}{\partial          [\bm{X}_{4}]_{b,c,d}} [\bm{X}_{4}]_{b,c,d}=L_{\bm{\theta}}-H_{\bm{\theta}},
\end{aligned}
\end{equation}
where $L_{\bm{\theta}}$ is Eq.\ref{eq:cross_ent}, $H_{\bm{\theta}}$ is the entropy of network output. Obviously, the cost sum is positively correlated to the loss, thus negatively correlated to the accuracy.
With $Y_{bn}$ denoting the output of $n$-th node from $b$-th image, 
\begin{equation}
\label{eq:cost_sum_expand}
\begin{aligned}
    C_{\bm{Z}}
    &=\frac{1}{B}\sum_{b,n}[-Y_{bn_b}+\frac{\exp( Y_{bn})}{\sum_{q}\exp(Y_{bq})}Y_{bn}].
\end{aligned}
\end{equation}
\textbf{Positive at low accuracy}:
Exploiting normalization and weight initialization, we have:
\begin{equation*}
\label{eq:cost_sum_init}
\begin{aligned}
\mathbb{E}_{\bm{\theta}_0}[C_{\bm{Z}}>0],
\end{aligned}
\end{equation*}
since $\mathbb{E}_{y_{1},y_{2} \sim \mathcal{N}(0,\cdot)}[y_{1}\exp{(y_{1}+y_{2}})]>0$.

\textbf{Negative at high accuracy}: With operation parameters updated towards convergence, the probability of $b$-th image being classified to the correct label $n_b$ increases towards 1. 
Since $Y_{bn_b} = \max{\{Y_{bn}\}}$, we have
\begin{equation*}
\label{eq:cost_sum_update_appx}
\begin{aligned}
    C_{\bm{Z}} \propto & \sum_{n}[-Y_{bn_b}+\frac{\exp( Y_{bn})}{\sum_{q}\exp(Y_{bq})}Y_{bn}] \\
    \leq &\sum_{n}[-Y_{bn_b}+\frac{\exp( Y_{bn})}{\sum_{q}\exp(Y_{bq})}Y_{bn_b}] = 0.
\end{aligned}
\end{equation*}

Derivation details can be found in Appx.\ref{app:proof_thm3}.
\end{proof}

\begin{figure}[t]
    \centering
    \includegraphics[width=3.14in]{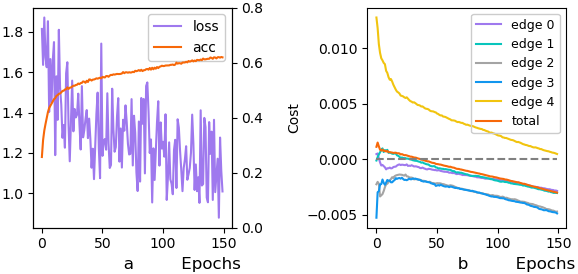}
    \caption{(a) Loss curve and accuracy curve. (b) Averaged cost on each edge and their sum. Note the correlation between (a) and (b).}
    \label{fig:cost_loss_acc}
    \vspace{-1em}
\end{figure}

As shown in Fig.\ref{fig:cost_loss_acc}, the cost of output edges at the last cell starts positive and eventually becomes negative. Intuitively, because ${\bm{\theta}}$ and $\bm{\alpha}$ are both optimized to minimize Eq.\ref{eq:objective}, training ${\bm{\theta}}$ is also minimizing the cost. More formally, Eq.\ref{eq:cost_ce_ent} bridges the dynamics of the cost sum with the learning dynamics \citep{saxe2014exact, liao2018dynamics} of ${\bm{\theta}}$ and ${\bm{\alpha}}$:
\begin{equation*}
\label{eq:cost_trend}
\begin{split}
    \frac{\partial C_{sum}(t)}{\partial t} &= \frac{\partial \mathbb{E}_{\bm{\alpha}}[C_{\bm{Z}}(t)]}{\partial t}= \frac{\partial \mathbb{E}_{\bm{\alpha}}[L_{\bm{\theta}}(t)]}{\partial t} - \frac{\partial \mathbb{E}_{\bm{\alpha}}[H_{\bm{\theta}}(t)]}{\partial t}.
\end{split}
\end{equation*}
The cost sum in one batch decreases locally if $\frac{\partial \mathbb{E}_{\bm{\alpha}}[L_{\bm{\theta}}(t)]}{\partial t} < \frac{\partial \mathbb{E}_{\bm{\alpha}}[H_{\bm{\theta}}(t)]}{\partial t}$ and increases locally if $\frac{\partial \mathbb{E}_{\bm{\alpha}}[L_{\bm{\theta}}(t)]}{\partial t} > \frac{\partial \mathbb{E}_{\bm{\alpha}}[H_{\bm{\theta}}(t)]}{\partial t}$. By Thm.\ref{thm:cost_dynamics}, the global trend of the cost sum at the last cell should be decreasing. 
\begin{remark} 
\label{thm:growth_trend}
    If cost at all edges are consistent with Thm.\ref{thm:cost_dynamics}, eventually the \textit{growing tendency} (\textbf{P1}) occurs.
\end{remark}

\subsection{Distinction of intermediate edges}
\label{sec:cost_of_inter_edge}

\begin{figure}[t]
    \centering
    \includegraphics[width=2.8in]{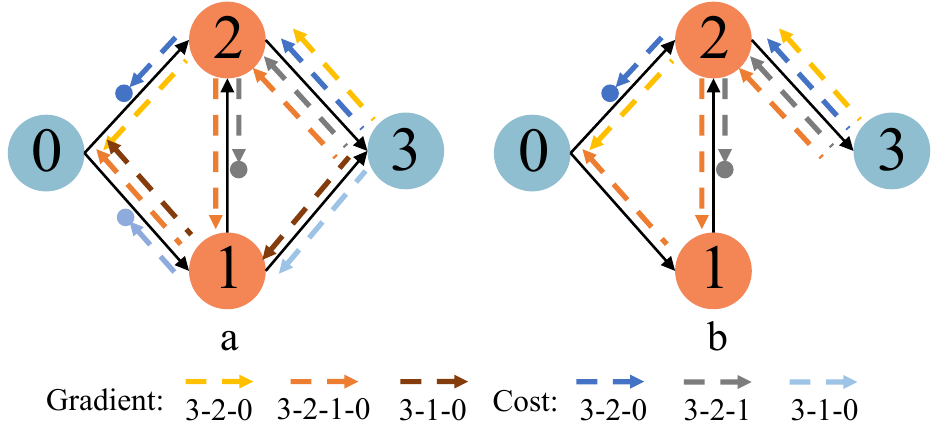}
    \caption{Gradient back-prop and cost assignment in (a) simplified cells and (b) modified cells. Note the difference on path 3-1-0.}
    \label{fig:simplified_modified_cell}
    \vspace{-1.5em}
\end{figure}

If all edges are born equally, then by the cost assignment mechanism in Sec.\ref{sec:cost_dist}, the edges finally picked would depend mainly on the task specification (dataset and objective) and randomness (lottery ticket hypothesis \citep{frankle2018lottery}), possibly also affected by the base architecture. But it does not explain the width preference. The width preference implies the distinction of intermediate edges. To analyze it, we further simplify the minimal cell assuming the symmetry of $X_{0}$ and $X_{1}$, as shown in Fig.\ref{fig:simplified_modified_cell}. 

\begin{figure}[t]
    \centering
    \includegraphics[width=3.2in]{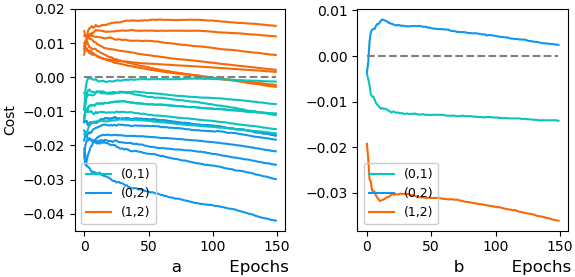}
    \vspace{-2mm}
    \caption{Averaged cost on each edge (a) in various settings on simplified cells and (b) fixing the operation on edge $(0,1)$ in modified cells. In (a) different curves with identical color differ in settings, see Appx.\ref{app:comp_details}.}
    \label{fig:compare_cost}
    \vspace{-1em}
\end{figure}

Fig.\ref{fig:compare_cost}(a) shows the cost estimated with the Monte Carlo estimation, averaged over operations. Cost at intermediate $edge (1,2)$ is higher than cost of the same operation at $edge (0,2)$, no matter whether there is sampling on $edge (0,1)$ or not, or how many operation candidates on each edge. See various settings in Appx.\ref{app:comp_details}.


We conjecture that the distinctiveness of the cost at intermediate edges is associated with the fact that it is less trained than input edges. It may be their lag in training that induces a lag in the cost-decreasing process, with a higher loss than frequently-trained counterparts. Why are they less trained? Note that in $\mathcal{A}$ every input must be followed by an output edge. Reflected in the simplified cell, \textit{${\bm{\theta}}_{0,1}$ and ${\bm{\theta}}_{0,2}$ are always trained as long as they are not sampled as \textit{None}}. Particularly, ${\bm{\theta}}_{0,1}$ is updated with gradients from two paths (3-2-1-0) and (3-1-0). When \textit{None} is sampled on $edge (1,2)$, ${\bm{\theta}}_{0,1}$ can be updated with gradient from path (3-1-0). However, \textit{when a \textit{None} is sampled on $edge (0,1)$, ${\bm{\theta}}_{1,2}$ cannot be updated because its input is zero}. Even if \textit{None} is not included in $edge (0,1)$, there are more model instances on path (3-2-1-0) than path (3-2-0) and (3-1-0) that share the training signal. 

We design a controlled experiment by deleting the output $edge(1,3)$ and fixing operation at $edge (0,1)$, as shown in Fig.\ref{fig:simplified_modified_cell}(b). This is no longer an instance of $\mathcal{A}$ since one intermediate node is not connected to the output node. But in this base architecture path (3-2-1-0) can be trained with an equal frequency as (3-2-0). As shown in Fig.\ref{fig:simplified_modified_cell}(d), the cost bias on $edge(1,2)$ is resolved. Interestingly, path (3-2-1-0) is deeper than (3-2-0), but the cost on $edge(0,2)$ becomes positive and dropped. \textit{The preference of the search algorithm is altered towards depth over width}. Hence the distinction of intermediate edges is likely due to unequal training between edges caused by base architecture design in $\mathcal{A}$, subnet sampling and weight sharing. 

\begin{remark}
\label{thm:inter_cost}
In $\mathcal{A}$, with subnet sampling and weight sharing, for edges pointing to the same node, cost at intermediate edges is higher than input edges.
\end{remark} 

Given a system in $\mathcal{A}$ following Remark \ref{thm:growth_trend}, the cost sum eventually decreases to become negative. \textit{Since the cost at intermediate edges tend to be higher than cost on input edges, if there is at least one edge with positive cost in the middle of evolution, this one is probably an intermediate edge}. It then becomes clear why intermediate edges recovers from the initial drop much later than input edges. 

\begin{remark}
In $\mathcal{A}$, with Remark \ref{thm:growth_trend}\&\ref{thm:inter_cost}, the \textit{width preference} \textbf{(P2)} occurs during the wiring evolution, even though sometimes intermediate edges finally recover.  
\end{remark}

\subsection{Effect of bi-level optimization}
\label{sec:bi_level_effect}

\begin{figure}[!tbp]
\vspace{-3.5mm}
\begin{minipage}{0.22\textwidth}
        \includegraphics[width=1.5in,height=1.34in]{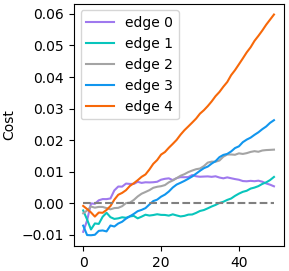}
\end{minipage}
\begin{minipage}{0.22\textwidth}
    \vspace{-3mm}
    \begin{tabular}{c|c|c}
    \hline
     & train & search \\
    \hline\hline
    $C_{correct}$ & -0.11 & -0.11 \\
    $L_{correct}$ & 0.06 & 0.08 \\
    $H_{correct}$ & 0.17 & 0.19 \\
    \hline
    $C_{wrong}$ & 0.80 & \textbf{2.18} \\
    $L_{wrong}$ & 1.70 & \textbf{2.92} \\
    $H_{wrong}$ & 0.90 & \textbf{0.74} \\
    \hline
    \end{tabular}
\end{minipage}
\caption{(a) Cost of edges in DARTS with bi-level optimization. y-axis for cost, x-axis for epochs. (b) Comparison of $C_{\bm{\theta}}$, $L_{\bm{\theta}}$ and $H_{\bm{\theta}}$ between the training set and the search set.}
\label{fig:darts_edge_cost}
\vspace{-1em}
\end{figure}

Until now we do not distinguish between single-level and bi-level optimization on $\bm{\alpha}$. This is because the cost minimization formulation Eq.\ref{eq:cost_obj} generally applies to both. However, that every edge drops and almost none of them finally recovers in DARTS's bi-level version seems exceptional. 
As shown in Fig.\ref{fig:darts_edge_cost}(a), the cost sum of edges even increases, which is to the opposite of Thm.\ref{thm:cost_dynamics} and Remark \ref{thm:inter_cost}.  Because the difference between bi-level optimization and the single-level is that the cost is calculated on a held-out search set, we scrutinize its classification loss $L_{\bm{\theta}}$ and output entropy $H_{\bm{\theta}}$, whose difference by Thm.\ref{thm:cost_dynamics} is the cost sum. 

Fig.\ref{fig:darts_edge_cost}(b) shows the comparison of $L_{\bm{\theta}}$ and $H_{\bm{\theta}}$ in the training set and the search set. For correct classification, $L_{\bm{\theta}}$ and $H_{\bm{\theta}}$ are almost comparable in the training set and the search set. But for data classified incorrectly, the classification loss $L_{\bm{\theta}}$ is much larger in the search set. That is, data in the search set are classified poorly. This can be be explained by overfitting. Apart from that, $H_{\bm{\theta}}$ in the search set is much lower than its counterpart in the training set. This discrepancy in $H_{\bm{\theta}}$ was previously discussed in the Bayesian deep learning literature: the softmax output of deep neural nets under-estimates the uncertainty. We would like to direct readers who are interested in a theoretical derivation to \citet{gal2016dropout}. \textit{Overall, subnetworks are erroneously confident in the held-out set, on which their larger $L_{\bm{\theta}}$ actually indicates their misclassification}. As a result, the cost sum in bi-level optimization becomes more and more positive. \textit{None} operation is chosen at all edges. 

\begin{remark}
\label{thm:bi-level}
The cost from the held-out set may incline towards positive values due to false classification and erroneously captured uncertainty. Hereby catastrophic failures (\textbf{P3}) occur in bi-level optimization.
\end{remark}

\section{Discussion}

In this work we study the underlying mechanism of wiring discovery in differentiable neural architecture search, motivated by some universal patterns in its evolution. We introduce a formulation that transfers NAS to a tractable local cost minimization problem, unifying existing frameworks. Since this cost is not static, 
we further investigate its assignment mechanism and learning dynamics. We discover some implicit inductive biases in existing frameworks, namely
\begin{itemize}[leftmargin=*,noitemsep,nolistsep]
    \item The cost assignment mechanism for architecture $\bm{\alpha}$ is non-trivially different from the credit assignment in gradient back-prop for parameters $\bm{\theta}$. Exaggerating this discrepancy, base architectures and training schemes diverge from the widely accepted assumption of facilitating the search on intermediate edges.  
    \item In training, cost decreases from positive and eventually turns negative, promoting the tendency of growth in topology. If this decreasing process is hindered or reversed, as in bi-level optimization, there will be a catastrophic failure in wiring discovery. 
\end{itemize}

To conclude, some topologies are chosen by existing differentiable NAS methods not because of being generally better, but because they fit these methods better. \textit{The observed regularity is a product of \textit{dedicated} base architectures and training schemes in these methods, rather than global optimality}. This conclusion calls for some deeper questions: Are there any ways to circumvent these biases without sacrificing generality? Are there any other implicit biases in existing NAS methods? Since this conclusion challenges some of the common assumptions in the community of Differentiable Neural Architecture Search, we want to make the following suggestions:
\begin{itemize}[leftmargin=*,noitemsep,nolistsep]
    \item Clearly state what is expected to be searched, \textit{i.e.} whether it is the operations or wiring topology or both. Frankly illustrate the evolution process according to the final architecture derivation schemes. 
    \item Investigate the implicit inductive biases imposed by normalization schemes in Differentiable NAS. 
    \item If weight sharing is necessary, investigate how sampling networks from base architectures with different topology affects the training of candidate operations. 
    \item Before we understand more about the implicit biases discovered in this work, be cautious when using bi-level optimization to search for wiring topology. 
\end{itemize}

Fortunately, efforts have been made to explore some of these questions \citep{liang2019darts+}. Towards this end, we hope our work can draw more attention to the theoretical study of Differentiable Neural Architecture Search.

\subsubsection*{Acknowledgements}
This work is mainly done at SenseTime Research Hong Kong. SH and XL are also partially supported by Hong Kong RGC GRF No. 14200220, Theme-based Research Scheme T45-407/19N, ITF grant No. ITS/254/19, and SHIAE grant No. MMT-p1-19. We also would like to thank the reviewers for their valuable comments and efforts towards improving our paper.


\bibliography{reference}
\bibliographystyle{apalike}






\newpage
\appendix

\onecolumn












\section{Evolution details of DARTS and DSNAS}
\label{app:evolution}
\subsection{DSNAS (single-level optimization)}
\begin{figure}[ht]
    \centering
    \includegraphics[width=3.0in]{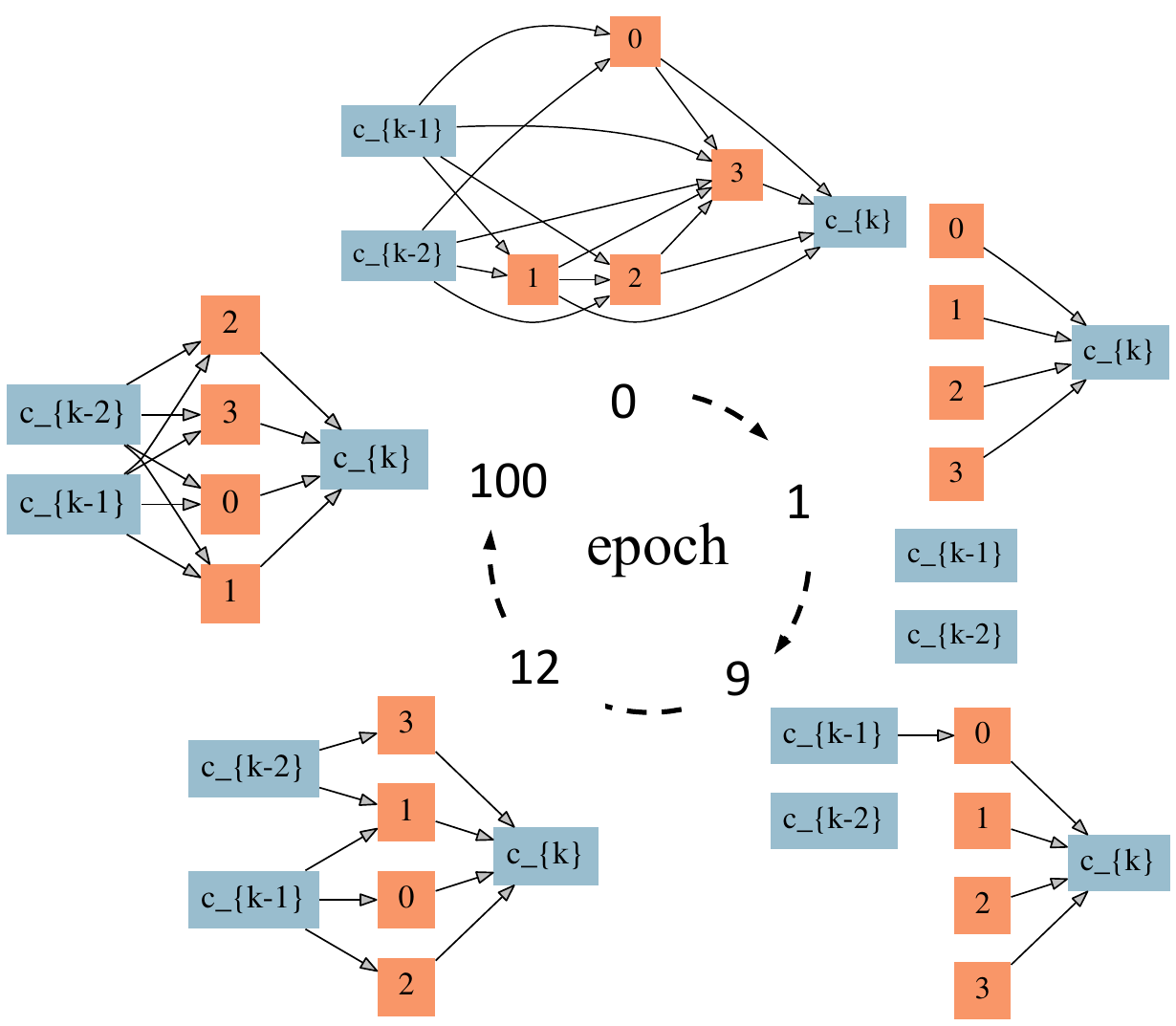}
    \caption{Evolution of the wiring topology with the largest probability in DSNAS. Edges are \textit{dropped} in the beginning. Final cells show preference towards width.}
    \label{fig:dsnas_evolve_appx}
\end{figure}

\begin{figure}[h]
    \centering
    \includegraphics[width=3.1in]{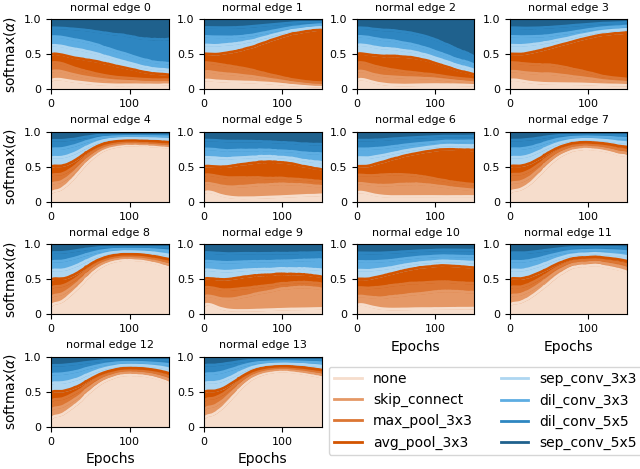}
    \caption{Evolution of $p_{\alpha}$ on each edge in DSNAS. 4, 7, 8, 11, 12, 13 are intermediate edges where \textit{None} dominates others. This domination is manifested as the \textit{width preference}. The rest are input edges, where \textit{None} only dominates in the beginning. Other operations then become the most likely ones, exhibiting the \textit{growth tendency}.}
    \label{fig:dsnas_update_alpha_appx}
\end{figure}

\subsection{DARTS (bi-level optimization)}
\begin{figure}[h]
    \centering
    \includegraphics[width=3.0in]{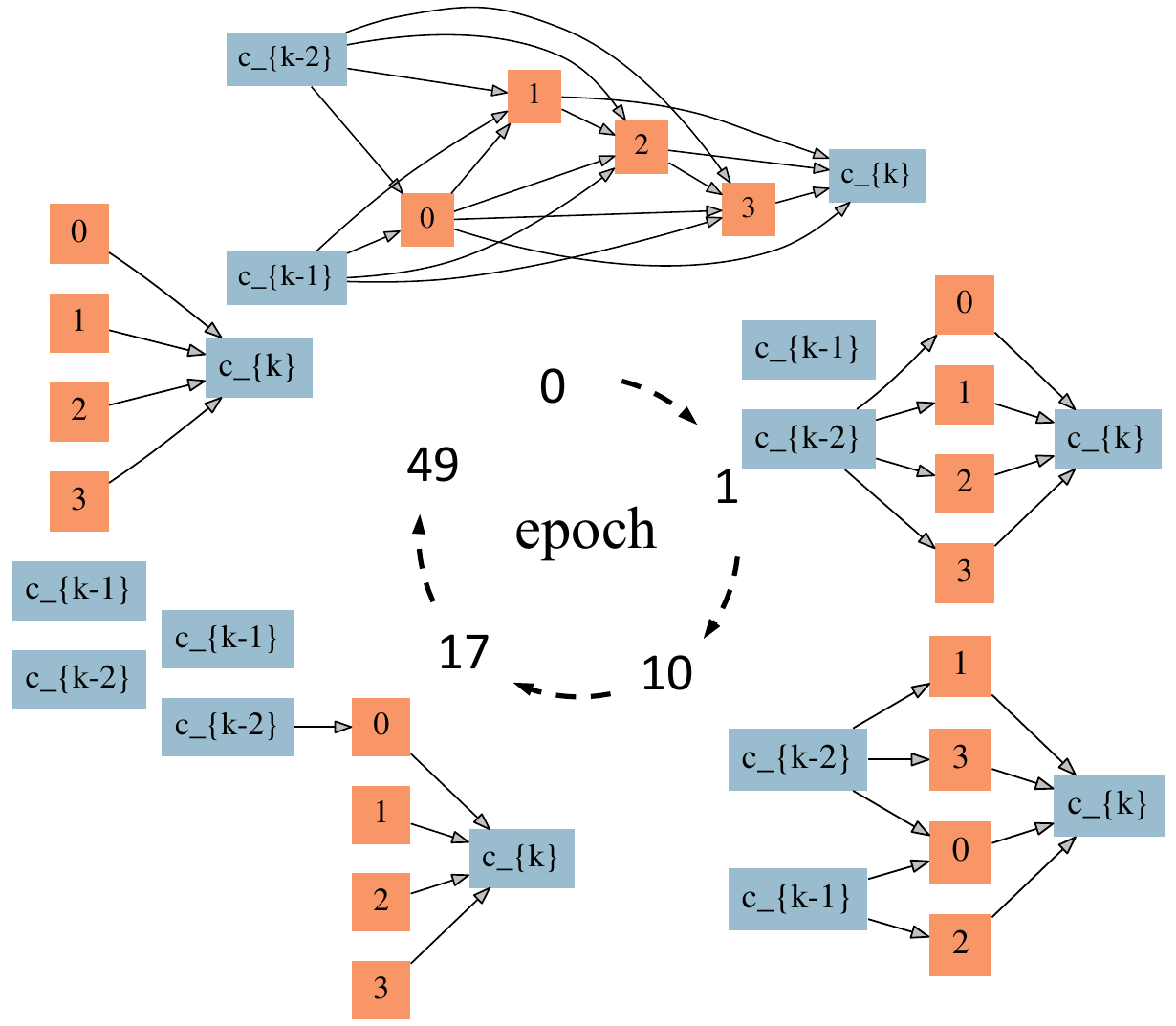}
    \caption{Evolution of the wiring topology with the largest probability in DARTS. Most edges are \textit{dropped} in the beginning, all edges are dropped after 50 epochs of training.}
    \label{fig:darts_evolve_appx}
\end{figure}

\begin{figure}[h]
    \centering
    \includegraphics[width=3.1in]{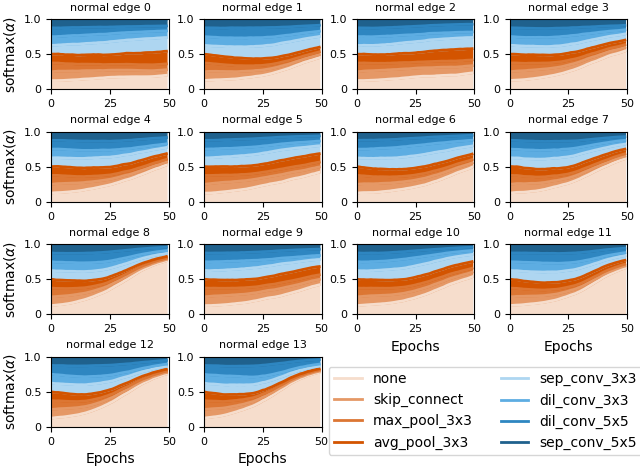}
    \caption{Evolution of $p_{\alpha}$ with epochs on each edge in DARTS with bi-level optimization. Different from those with single-level optimization, all edges are gradually dropped with training going on. At the end of the evolution, \textit{None} has the largest probability on all edges.}
    \label{fig:darts_update_alpha_appx}
\end{figure}
\newpage

\section{Unifying differentiable NAS}
\label{app:cost_darts_prox}
\subsection{DARTS}
Different from SNAS, DARTS \citep{liu2018darts} utilizes an attention mechanism to construct the parent network. To align with (\ref{eq:cost_obj}), we show that with $\hat{\bm{Z}}_{i,j}^{k}=\frac{\exp(\bm{\alpha}_{i,j}^{k})}{\sum_{m}\exp(\bm{\alpha}_{i,j}^{m})}$: 
\begin{equation*}
\begin{split}
        \frac{\partial \mathcal{L}}{\partial \bm{\alpha}_{i,j}^{}} &= \frac{\partial \mathcal{L}}{\partial x_{j}}\frac{\partial x_{j}}{\partial \bm{\alpha}_{i,j}^{}} = [\frac{\partial \mathcal{L}}{\partial x_{j}}\bm{O}^{}_{i, j}(x_{i})]_{c}\frac{\partial \hat{\bm{Z}}_{i,j}^{}}{\partial \bm{\alpha}_{i,j}^{}} \\&= [\frac{\partial \mathcal{L}}{\partial \hat{\bm{Z}}_{i,j}^{}}]_{c}\frac{\partial \hat{\bm{Z}}_{i,j}^{}}{\partial \bm{\alpha}_{i,j}^{}} = \frac{\partial \hat{\bm{Z}}_{i,j}^{}}{\partial \bm{\alpha}_{i,j}^{}}C(\hat{\bm{Z}}_{i,j}^{}),
\end{split}
\end{equation*}
where as in (\ref{eq:cost_obj_cont_snas}) $[\cdot]_{c}$ denotes $\cdot$ is a cost function independent from the gradient calculation \textit{w.r.t.} $\bm{\alpha}$.  

\subsection{ProxylessNAS}
ProxylessNAS \citep{cai2018proxylessnas} inherits DARTS's learning objective, and introduce the BinnaryConnect technique to empirically save memory. To achieve that, they propose the following approximation:
\begin{equation*}
    \frac{\partial \mathcal{L}}{\partial \bm{\alpha}_{i,j}^{}} = \frac{\partial \mathcal{L}}{\partial \hat{\bm{Z}}_{i,j}}\frac{\partial \hat{\bm{Z}}_{i,j}}{\partial \bm{\alpha}_{i,j}^{}} \approx \sum_{k}[\frac{\partial \mathcal{L}}{\partial {{Z}}_{i,j}^{k}}]_{c}\frac{\partial \hat{{Z}}_{i,j}^{k}}{\partial \bm{\alpha}_{i,j}^{}}  = \frac{\partial \hat{\bm{Z}}_{i,j}^{}}{\partial \bm{\alpha}_{i,j}^{}}C({\bm{Z}}_{i,j}^{}),
\end{equation*}
where as in (\ref{eq:cost_obj_cont_snas}) $[\cdot]_{c}$ denotes $\cdot$ is a constant for the gradient calculation \textit{w.r.t.} $\bm{\alpha}$. Obviously, it is also consistent with (\ref{eq:cost_obj}). 

\section{Cost mean when stacking 8 cells in DSNAS}
\label{app:cost_8_cells}

\begin{figure}[h]
    \centering
    \includegraphics[width=3.3in]{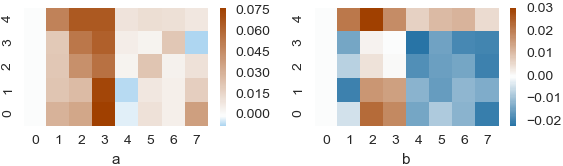}
    \caption{Cost mean statistics for each operation (x-axis) on each edge (y-axis) (a) at initialization and (b) near $\theta$'s convergence of by stacking 8 minimal cells. Operation 0: none, Operation 1: skip connect, Operation 2: max\_pool\_3x3, Operation 3: avg\_pool\_3x3, Operation 4: sep\_conv\_3x3, Operation 5: dil\_conv\_3x3, Operation 6: dil\_conv\_5x5, Operation 7: sep\_conv\_5x5.}
    \label{fig:cost_mean_minimal_8layer_appx}
\end{figure}

\begin{figure}[h]
    \centering
    \includegraphics[width=3.3in]{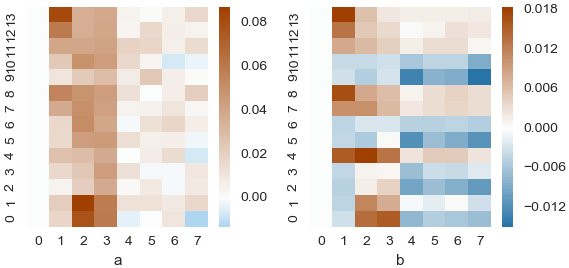}
    \caption{Cost mean statistics for each operation (x-axis) on each edge (y-axis) (a) at initialization and (b) near $\theta$'s convergence by stacking 8 original cells as in DARTS and SNAS. Operation 0: none, Operation 1: skip connect, Operation 2: max\_pool\_3x3, Operation 3: avg\_pool\_3x3, Operation 4: sep\_conv\_3x3, Operation 5: dil\_conv\_3x3, Operation 6: dil\_conv\_5x5, Operation 7: sep\_conv\_5x5.}
    \label{fig:cost_mean_original_8layer_appx}
\end{figure}

\section{Proof of Thm. \ref{thm:bn0}}
\label{app:proof_thm2}

\begin{figure}[h]
    \centering
    \includegraphics[width=2.6in]{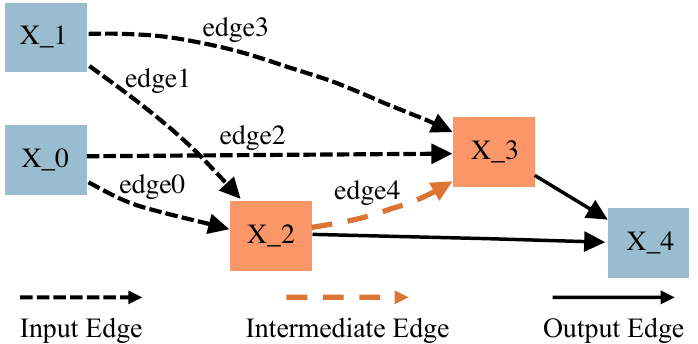}
    \caption{A minimal cell is a cell with 2 \textit{intermediate nodes} (orange), two \textit{input nodes} and one \textit{output node} (blue). Edges connecting input nodes and intermediate nodes ($edge 0-edge3$) are called \textit{input edges}; edges between two intermediate nodes are \textit{intermediate edges} ($edge 4$); others are \textit{output edges} which are skip-connections and concatenated together.}
    \label{fig:minimal_cell_apx}
\end{figure}

\begin{theorem} 
\label{thm:bn0_app}
A path does not distribute cost from its output edge after passing one intermediate edge.
\begin{equation*}
\label{eq:cost_edge_01_sec_apx}
\begin{aligned}
    C(Z_{0,2}^{s})=\frac{\partial L_{\bm{\theta}}}{\partial \bm{X}_{3}^{2}}\frac{\partial \bm{X}_{3}^{2}}{\partial \bm{X}_{2}^{0}}\bm{X}_{2}^{0}=0.
\end{aligned}
\end{equation*}
\end{theorem}

\begin{proof}
For each intermediate edge, three operations (ReLU, Conv/pooling, BN) are sequentially ordered as shown in Fig. \ref{fig:op_bn_edge4_appx}. To prove Thm.~\ref{thm:bn0}, we analyse the effect of each operation on the cost assignment in the intermediate edge. Let $\bm{X}_{2}^{0}\in \mathbb{R}^{B\times C_{in} \times W_{in} \times H_{in}}$ denotes the input of Conv operation on edge $4$\footnote{We do not consider ReLU operation (before Conv) here, since ReLU operation obviously satisfies the proof. Thm. \ref{thm:bn0} can be easily generalized to pooling operations.}, $\bm{X}\in \mathbb{R}^{B\times C_{out} \times W_{out} \times H_{out}}$ denotes the Conv operation output, $W\in \mathbb{R}^{C_{out}\times C_{in}\times K\times K}$ is the filter weight in the Conv operation. The full proof consists of three following steps.

\begin{figure}[h!]
    \centering
    \includegraphics[width=2.8in]{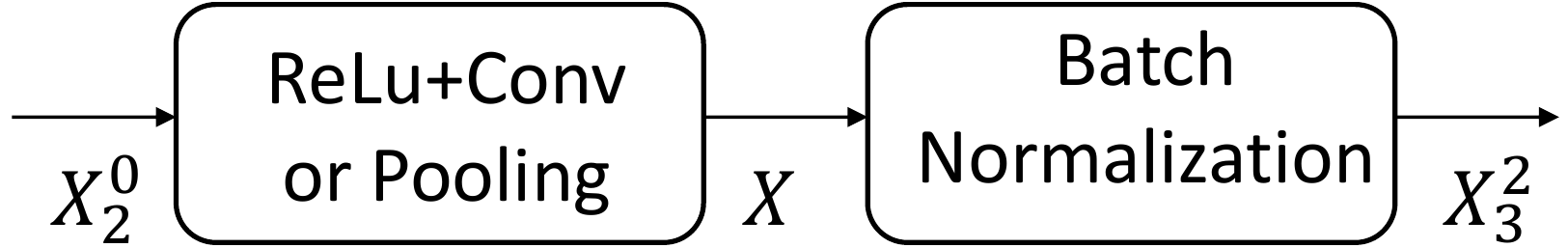}
    \caption{Stacking order of ReLU, Conv/pooling and BN operations on edge $4$.}
    \label{fig:op_bn_edge4_appx}
\end{figure}

\textbf{Step 1}: By expanding $C(Z_{0,2}^{s})$ at path (4-3-2-0), we have:
\begin{equation} 
\label{eq:cost_edge_02_sec_apx}
    \frac{\partial L_{\bm{\theta}}}{\partial \bm{X}_{3}^{2}}\frac{\partial \bm{X}_{3}^{2}}{\partial \bm{X}_{2}^{0}}\bm{X}_{2}^{0}=\frac{\partial L_{\bm{\theta}}}{\partial \bm{X}_{3}^{2}} \frac{\partial \bm{X}_{3}^{2}}{\partial \bm{X}} \frac{\partial \bm{X}}{\partial \bm{X}_{2}^{0}}\bm{X}_{2}^{0}.
\end{equation}

\textbf{Step 2}: Utilizing the \textbf{linear property} of Conv operation, we can simplify the above equation as:
\begin{equation}
\label{eq:cost_edge_02_conv_apx}
    \underline{\frac{\partial L_{\bm{\theta}}}{\partial \bm{X}_{3}^{2}} \frac{\partial \bm{X}_{3}^{2}}{\partial \bm{X}} \frac{\partial \bm{X}}{\partial \bm{X}_{2}^{0}}}\bm{X}_{2}^{0}=\frac{\partial L_{\bm{\theta}}}{\partial \bm{X}_{3}^{2}} \frac{\partial \bm{X}_{3}^{2}}{\partial \bm{X}} \bm{X}.
\end{equation}

Proof of Step 2: To derive Eq.(\ref{eq:cost_edge_02_conv_apx}), we first calculate the gradient w.r.t Conv input $\bm{X}_{2}^{0}$ (\textbf{black-underlined part}). Note that $w_{in}$ ($w_{out}$), $h_{in}$ ($h_{out}$), $c_{in}$ ($c_{out}$) and $b$ denote the index of width, height, channel and batch dimensions of Conv input (output), and $K$ is the Conv filter width and height.
\begin{equation}
\label{eq:conv_backward_apx}
\begin{aligned}
    &\left[\underline{\frac{\partial L_{\bm{\theta}}}{\partial \bm{X}_{3}^{2}} \frac{\partial \bm{X}_{3}^{2}}{\partial \bm{X}} \frac{\partial \bm{X}}{\partial \bm{X}_{2}^{0}}}\right]_{b c_{in} w_{in} h_{in}}=\sum_{c_{out}}\sum_{\substack{w_{out}=w_{in}-K\\ h_{out}=h_{in}-K}}^{\substack{w_{in} \\ h_{in}}} \frac{\partial L_{\bm{\theta}}}{\bm{X}_{b c_{out} w_{out} h_{out}}}W_{c_{out}c_{in}(w_{in}-w_{out})(h_{in}-h_{out})}
\end{aligned}
\end{equation}

Based on Eq.(\ref{eq:conv_backward_apx}), we can simplify the left part of Eq.(\ref{eq:cost_edge_02_conv_apx}) by doing an \textbf{element-wise calculation} on the width ($w$), height ($h$), channel ($c$) and batch ($b$) dimensions.
\begin{equation}
\label{eq:cost_conv_apx}
\begin{aligned}
    &\underline{\frac{\partial L_{\bm{\theta}}}{\partial \bm{X}_{3}^{2}} \frac{\partial \bm{X}_{3}^{2}}{\partial \bm{X}} \frac{\partial \bm{X}}{\partial \bm{X}_{2}^{0}}}\bm{X}_{2}^{0}\\
    &=\sum_{b,c_{out},w_{in},h_{in}}\ \sum_{w_{out}=w_{in}-K}^{w_{in}}\ \sum_{h_{out}=h_{in}-K}^{h_{in}}\ \frac{\partial L_{\bm{\theta}}}{\partial \bm{X}_{b c_{out} w_{out} h_{out}}}\sum_{c_{in}}W_{c_{out}c_{in}(w_{in}-w_{out})(h_{in}-h_{out})}[\bm{X}_{2}^{0}]_{b c_{in}w_{in}h_{in}}\\
    &=\sum_{b,c_{out},w_{out},h_{out}}\ \sum_{w_{in}=w_{out}}^{w_{out}+K}\ \sum_{h_{in}=h_{out}}^{h_{out}+K}\ \frac{\partial L_{\bm{\theta}}}{\partial \bm{X}_{b c_{out} w_{out} h_{out}}}\sum_{c_{in}}W_{c_{out}c_{in}(w_{in}-w_{out})(h_{in}-h_{out})}[\bm{X}_{2}^{0}]_{b c_{in}w_{in}h_{in}}\\
    &=\sum_{b,c_{out},w_{out},h_{out}}\ \frac{\partial L_{\bm{\theta}}}{\partial \bm{X}_{b c_{out} w_{out} h_{out}}}\mathunderline{red}{\sum_{w_{in}=w_{out}}^{w_{out}+K}\ \sum_{h_{in}=h_{out}}^{h_{out}+K}}\mathunderline{red}{\sum_{c_{in}}W_{c_{out}c_{in}(w_{in}-w_{out})(h_{in}-h_{out})}[\bm{X}_{2}^{0}]_{b c_{in}w_{in}h_{in}}}\\    
    &=\sum_{b,c_{out},w_{out},h_{out}}\frac{\partial L_{\bm{\theta}}}{\partial \bm{X}_{b c_{out} w_{out} h_{out}}} \mathunderline{red}{\bm{X}_{b c_{out}w_{out}h_{out}}}\\
    &=\frac{\partial L_{\bm{\theta}}}{\partial \bm{X}_{3}^{2}} \frac{\partial \bm{X}_{3}^{2}}{\partial \bm{X}} \bm{X},
\end{aligned}
\end{equation}
where the second equality holds by changing the order of indexes $w_{in}$, $h_{in}$, $w_{out}$, $h_{out}$. Using the linear property of Conv operation, the \textcolor{red}{\textbf{red-underlined}} part in the forth equality can be derived due to:
\begin{equation}
\label{eq:conv_forward}
\begin{aligned}
    &{\bm{X}}_{b c_{out}w_{out}h_{out}}=\sum_{c_{in}} \sum_{\substack{w_{in}=w_{out} \\ h_{in}=h_{out}}}^{\substack{w_{out}+K \\ h_{out}+K }} W_{c_{out}c_{in}(w_{in}-w_{out})(h_{in}-h_{out})}{[\bm{X}_{2}^{0}]}_{b c_{in} w_{in} h_{in}}.
\end{aligned}
\end{equation}

\textbf{Step 3}: We have shown Conv/pooling operations does not change the cost value in the intermediate edge, so next we analyse the effect of batch normalization~\citep{ioffe2015batch} on the cost assignment. Exploiting the \textbf{property of batch normalization}~\citep{ioffe2015batch}, we have:
\begin{equation}
\label{eq:cost_edge_02_bn_apx}
\begin{aligned}
    \underline{\frac{\partial L_{\bm{\theta}}}{\partial \bm{X}_{3}^{2}} \frac{\partial \bm{X}_{3}^{2}}{\partial \bm{X}}} \bm{X}=\underline{\frac{\partial L_{\bm{\theta}}}{\partial \bm{X}}} \bm{X}=0.
\end{aligned}
\end{equation}

Proof of Step 3: Similar as step2, we compute the gradient w.r.t. the Batch Normalization input $\bm{X}$ (\textbf{black-underlined part}). Before we start, we first show the batch normalization process as below:
\begin{equation}
\label{eq:bn_forward_apx}
\begin{aligned}
    &\mu_{c} = \frac{1}{BD}\sum_{b,d}\bm{X}_{b,c,d}, 
    &\sigma_{c}^{2} = \frac{1}{BD}\sum_{b,d}\left(\bm{X}_{b,c,d}-\mu_c\right)^2,\\
    &[\hat{\bm{X}}^{2}_{3}]_{b,c,d} = \frac{\bm{X}_{b,c,d}-\mu_c}{\sqrt{\sigma_{c}^{2}}}, 
    &[\bm{X}^{2}_{3}]_{b,c,d}=\gamma_c*[\hat{\bm{X}}^{2}_{3}]_{b,c,d}+\beta_c.
\end{aligned}
\end{equation}
where $\bm{X}$ and $\bm{X}^{2}_{3}$ denote the input and output of BN operation, $\mu_{c}$ and $\sigma_{c}^{2}$ are the mean and variance statistics of $c$-th channel. Note that d denotes the spatial size $Width\times Height$.
Then, the gradients with respect with the output $\bm{X}_{b,c,d}$ (\textbf{black-underlined part}) are calculated based on chain rule:
\begin{equation}
\label{eq:bn_output_backward_apx}
\begin{aligned}
    \underline{\frac{\partial L_{\bm{\theta}}}{\partial \bm{X}_{b,c,d}}}&=\frac{\partial L_{\bm{\theta}}}{\partial [\bm{X}^{2}_{3}]_{b,c,d}}\frac{\partial [\bm{X}^{2}_{3}]_{b,c,d}}{\partial \bm{X}_{b,c,d}}+\textcolor{blue}{\frac{\partial L_{\bm{\theta}}}{\partial \mu_{c}}}\frac{\partial \mu_{c}}{\partial \bm{X}_{b,c,d}}+\textcolor{magenta}{\frac{\partial L_{\bm{\theta}}}{\partial \sigma_{c}^{2}}}\frac{\partial \sigma_{c}^{2}}{\partial \bm{X}_{b,c,d}}(\bm{chain\ rule})\\
    &=\frac{\partial L_{\bm{\theta}}}{\partial [\bm{X}^{2}_{3}]_{b,c,d}}\frac{\gamma_c}{\sqrt{\sigma_c^2}}+\textcolor{blue}{\frac{\partial L_{\bm{\theta}}}{\partial \mu_{c}}}\frac{1}{BD}+\textcolor{magenta}{\frac{\partial L_{\bm{\theta}}}{\partial \sigma_{c}^{2}}}\frac{2(\bm{X}_{b,c,d}-\mu_c)}{BD}\\
    &=\frac{\partial L_{\bm{\theta}}}{\partial [\bm{X}^{2}_{3}]_{b,c,d}}\frac{\gamma_c}{\sqrt{\sigma_c^2}}-\frac{1}{BD}\sum_{b,d}\frac{\partial L_{\bm{\theta}}}{\partial [\bm{X}^{2}_{3}]_{b,c,d}}\frac{\gamma_c}{\sqrt{\sigma_c^2}}-\frac{\gamma_c}{\sqrt{\sigma_c^2}}(\sum_{b,d}\frac{\partial L_{\bm{\theta}}}{\partial [\bm{X}^{2}_{3}]_{b,c,d}}[\hat{\bm{X}}^{2}_{3}]_{b,c,d})\frac{[\hat{\bm{X}}^{2}_{3}]_{b,c,d}}{CD}.
\end{aligned}
\end{equation}
where the gradients with respect to the mean $\mu_{c}$ ($\textcolor{blue}{\frac{\partial L_{\bm{\theta}}}{\partial \mu_{c}}}$) and variance $\sigma_{c}^{2}$ ($\textcolor{magenta}{\frac{\partial L_{\bm{\theta}}}{\partial \sigma_{c}^{2}}}$) are shown as below:
\begin{equation}
\label{eq:bn_mean_sigma_backward_apx}
\begin{aligned}
    \textcolor{blue}{\frac{\partial L_{\bm{\theta}}}{\partial \mu_{c}}}&=\sum_{b,d}\frac{\partial L_{\bm{\theta}}}{\partial [\bm{X}^{2}_{3}]_{b,c,d}}\frac{\partial [\bm{X}^{2}_{3}]_{b,c,d}}{\partial \mu_{c}}-\frac{\partial L_{\bm{\theta}}}{\partial \sigma_{c}^{2}}\frac{\partial \sigma_{c}^{2}}{\partial \mu_{c}} (\bm{chain\ rule})\\
    &=-\sum_{b,d}\frac{\partial L_{\bm{\theta}}}{\partial [\bm{X}^{2}_{3}]_{b,c,d}}\frac{\gamma_c}{\sqrt{\sigma_c^2}}-\textcolor{orange}{\sum_{b,d}\frac{\partial L_{\bm{\theta}}}{\partial \sigma_{c}^{2}}\frac{X_{b,c,d}-\mu_c}{BD/2}}\\
    &=-\sum_{b,d}\frac{\partial L_{\bm{\theta}}}{\partial [\bm{X}^{2}_{3}]_{b,c,d}}\frac{\gamma_c}{\sqrt{\sigma_c^2}}\\
    \textcolor{magenta}{\frac{\partial L_{\bm{\theta}}}{\partial \sigma_{c}^{2}}}&=\sum_{b,d}\frac{\partial L_{\bm{\theta}}}{\partial [\bm{X}^{2}_{3}]_{b,c,d}}\frac{\partial [\bm{X}^{2}_{3}]_{b,c,d}}{\partial \sigma_{c}^{2}} (\bm{chain\ rule})\\
    &=-\frac{\gamma_c}{2}\sum_{b,d}\frac{\partial L_{\bm{\theta}}}{\partial [\bm{X}^{2}_{3}]_{b,c,d}}\frac{X_{b,c,d}-\mu_{c}}{(\sigma_c^2)^{3/2}}\\
    &=-\frac{\gamma_c}{2}\sum_{b,d}\frac{\partial \mathcal{L}}{\partial [\bm{X}^{2}_{3}]_{b,c,d}}\frac{[\hat{\bm{X}}^{2}_{3}]_{b,c,d}}{\sigma_c^2}
\end{aligned}
\end{equation}
where the \textcolor{orange}{\textbf{orange-colored}} term is strictly 0 due to the \textbf{standard Gaussian mean property} $\sum_{b,d}\frac{[\hat{\bm{X}}^{2}_{3}]_{b,c,d}}{BD}=0$.

After getting the specific form of the \textbf{black-underlined part}, we can directly derive Eq.(\ref{eq:cost_edge_02_bn_apx}) by doing an \textbf{element-wise calculation}:
\begin{equation}
\label{eq:cost_bn_cost_apx}
\begin{aligned}
    &\underline{\frac{\partial L_{\bm{\theta}}}{\partial \bm{X}_{3}^{2}} \frac{\partial \bm{X}_{3}^{2}}{\partial \bm{X}}}\bm{X}=\sum_{b,c,d}\frac{\partial L_{\bm{\theta}}}{\partial [\bm{X}^{2}_{3}]_{b,c,d}}\frac{\gamma_c\bm{X}_{b,c,d}}{\sqrt{\sigma_c^2}}-\frac{1}{BD}\sum_{b,c,d}(\sum_{b,d}\frac{\partial L_{\bm{\theta}}}{\partial [\bm{X}^{2}_{3}]_{b,c,d}})\frac{\gamma_c \bm{X}_{b,c,d}}{\sqrt{\sigma_c^2}}\\
    &-\sum_{b,c,d}\frac{\gamma_c}{BD}(\sum_{b,d}\frac{\partial L_{\bm{\theta}}}{\partial [\bm{X}^{2}_{3}]_{b,c,d}}[\hat{\bm{X}}^{2}_{3}]_{b,c,d})[\hat{\bm{X}}^{2}_{3}]_{b,c,d}\frac{\bm{X}_{b,c,d}}{\sqrt{\sigma_c^2}}\\
    &=\sum_{b,c,d}\frac{\partial L_{\bm{\theta}}}{\partial [\bm{X}^{2}_{3}]_{b,c,d}}\frac{\gamma_c(\bm{X}_{b,c,d}-\mu_c)}{\sqrt{\sigma_c^2}}-\mathunderline{orange}{\frac{1}{BD}\sum_{b,c,d}(\sum_{b,d}\frac{\partial L_{\bm{\theta}}}{\partial [\bm{X}^{2}_{3}]_{b,c,d}})\frac{\gamma_c(\bm{X}_{b,c,d}-\mu_c)}{\sqrt{\sigma_c^2}}}\\
    &-\sum_{b,c,d}\frac{\gamma_c}{BD}(\sum_{b,d}\frac{\partial L_{\bm{\theta}}}{\partial [\bm{X}^{2}_{3}]_{b,c,d}}[\hat{\bm{X}}^{2}_{3}]_{b,c,d})[\hat{\bm{X}}^{2}_{3}]_{b,c,d}\frac{\bm{X}_{b,c,d}}{\sqrt{\sigma_c^2}}\\
    &=\sum_{b,c,d}\frac{\partial L_{\bm{\theta}}}{\partial [\bm{X}^{2}_{3}]_{b,c,d}}\gamma_c[\hat{\bm{X}}^{2}_{3}]_{b,c,d}-\sum_{b,c,d}\frac{\gamma_c}{BD}(\sum_{b,d}\frac{\partial L_{\bm{\theta}}}{\partial [\bm{X}^{2}_{3}]_{b,c,d}}[\hat{\bm{X}}^{2}_{3}]_{b,c,d})[\hat{\bm{X}}^{2}_{3}]_{b,c,d}\frac{\bm{X}_{b,c,d}}{\sqrt{\sigma_c^2}}\\
    &=\sum_{c}\gamma_c(\sum_{b,d}\frac{\partial L_{\bm{\theta}}}{\partial [\bm{X}^{2}_{3}]_{b,c,d}}[\hat{\bm{X}}^{2}_{3}]_{b,c,d})(1-\sum_{b,d}\frac{[\hat{\bm{X}}^{2}_{3}]_{b,c,d}^2}{BD})\\
    &=0,
\end{aligned}
\end{equation}
where the \textcolor{orange}{\textbf{orange-underlined}} term is strictly 0 due to the \textbf{standard Gaussian mean property} $\sum_{b,d}\frac{[\hat{\bm{X}}^{2}_{3}]_{b,c,d}}{BD}=0$, the last equality also holds due to \textbf{standard Gaussian variance property} $\sum_{b,d}\frac{[\hat{\bm{X}}^{2}_{3}]^2_{b,c,d}}{BD} = 1$.

This result is consistent with \citep{tian2018theoretical}, in which batch normalization is proved to project the input gradient to the orthogonal complement space spanned by the batch normalization input and vector $\bm{1}$. Note that this result can be generalized to arbitrary back-propagation path involving intermediate edges and \textbf{other normalization methods}, like instance normalization \citep{ulyanov2016instance}, layer normalization \citep{ba2016layer}. See Appx.\ref{app:other_norm}.

One exception from the proof above is the \textit{skip} operation, in which no BN is added. 
Let $\bm{X}_{2}^{0}\in \mathbb{R}^{B\times C_{in} \times W_{in} \times H_{in}}$ denotes the input of skip operation on edge $4$. We have 
\begin{equation}
    \frac{\partial L_{\bm{\theta}}}{\partial \bm{X}_{3}^{2}} \frac{\partial \bm{X}_{3}^{2}}{\partial \bm{X}_{2}^{0}}\bm{X}_{2}^{0}=\frac{\partial L_{\bm{\theta}}}{\partial \bm{X}_{3}^{2}} \bm{X}_{3}^{2}.
\end{equation}
where $\bm{X}_{3}^{2}=\bm{X}_{2}^{0}$. Obviously, if \textit{skip} is sampled on edge 4, it does not block the cost assignment as in Thm \ref{thm:bn0}. However, the influence of the \textit{skip} operation on edge 4 can be ignored since the cost magnitude is much smaller than other operations.
\end{proof}

\section{Validation of Cor. 1.2}
\label{app:val_cor}
\begin{figure}[h!]
    \centering
    \includegraphics[width=3.2in]{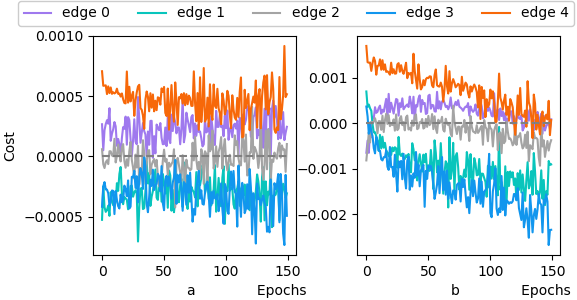}
    \caption{(a) Averaged cost on each edge of the second last cell. (b) Averaged cost on each edge of the last cell.The cost sum of edges in (a) is 0, but it is not 0 in (b). }
    \label{fig:compare_cost_last_two}
\end{figure}

\section{Proof of Thm. \ref{thm:cost_dynamics}}
\label{app:proof_thm3}
\begin{theorem}
\label{thm:cost_dynamics_app}
Cost at output edges of the last cell has the form $C_{\bm{Z}}=L_{\bm{\theta}}-H_{\bm{\theta}}$. 
It is negatively related to classification accuracy. It tends to be positive at low accuracy, negative at high accuracy.
\end{theorem}

\begin{figure}[h]
    \centering
    \includegraphics[width=3.2in]{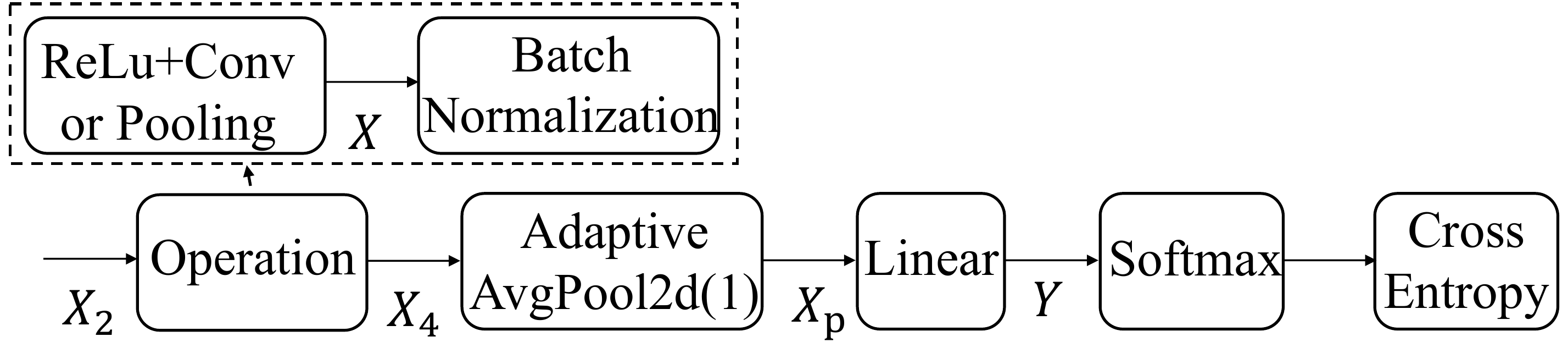}
    \caption{Post-processing the output from the last cell for loss.}
    \label{fig:last_cell_appx}
\end{figure}

\begin{proof} \textbf{Negatively related}:
We can first prove that for the last cell's output edges cost of one batch $M$ with sampled architecture $\bm{Z}$ has an equivalent form:
\begin{equation}
\label{eq:cost_ce_ent_apx}
\begin{aligned}
    \textcolor{red}{C_{\bm{Z}}}&=\sum_{b,c,d}\underline{\frac{\partial L_{\bm{\theta}}}{\partial          [\bm{X}_{4}]_{b,c,d}}} [\bm{X}_{4}]_{b,c,d}\\
    &=\frac{1}{BD}\sum_{b,c,d,n}\underline{[-w_{cn_b}+\frac{\exp( Y_{bn})}{\sum_{q}\exp(Y_{bq})}w_{cn}]}[\bm{X}_{4}]_{b,c,d}\\
    &=\frac{1}{B}\sum_{b,c,n}[-w_{cn_b}+\frac{\exp( Y_{bn})}{\sum_{q}\exp(Y_{bq})}w_{cn}][\bm{X}_{p}]_{b,c}\\
    &=\textcolor{red}{\frac{1}{B}\sum_{b,n}[-Y_{bn_b}+\frac{\exp( Y_{bn})}{\sum_{q}\exp(Y_{bq})}Y_{bn}]}\\
    &=\frac{1}{B}\sum_{b,n}[-\log(\exp(Y_{bn_b}))+\frac{\exp( Y_{bn})}{\sum_{q}\exp(Y_{bq})}Y_{bn}+ \log(\sum_{q}\exp(Y_{bq}))-\log(\sum_{q}\exp(Y_{bq}))]\\
    &=\frac{1}{B}\sum_{b}[-\log\frac{\exp( Y_{bn_b})}{\sum_{q}\exp(Y_{bq})}+\sum_{n}\frac{\exp( Y_{bn})}{\sum_{q}\exp(Y_{bq})}\log\frac{\exp( Y_{bn})}{\sum_{q}\exp(Y_{bq})}]\\
    &=L_{\bm{\theta}}-H_{\bm{\theta}},
\end{aligned}
\end{equation}
where $n_b$ is the corresponding $b$-th image class label, $L_{\bm{\theta}}=-\frac{1}{B}\log{\frac{\exp( Y_{bn_b})}{\sum_{q}\exp(Y_{bq})}}$ is Eq.\ref{eq:cross_ent}, $H_{\bm{\theta}}$ is the entropy of network output, $w_{ch}$ is the weight parameter in the linear layer, $[\bm{X}_{p}]_{b,c}=\frac{1}{D}\sum_{d}[\bm{X}_{4}]_{b,c,d}$ is an element of the output from the adaptive average pooling, $Y_{bn}=\sum_{c}[\bm{X}_{p}]_{b,c}w_{cn}$ is an element of the output of the last linear layer. Obviously, the cost sum is positively correlated to the loss, thus negatively correlated to the accuracy. The \textbf{black-underlined part} is derived by chain rule:
\begin{align*}
    \underline{\frac{\partial L_{\bm{\theta}}}{\partial[\bm{X}_{4}]_{b,c,d}}}&=\sum_{n}\frac{\partial L_{\bm{\theta}}}{\partial Y_{bn}} \frac{\partial Y_{bn}}{\partial [\bm{X}_{p}]_{b,c}}\frac{\partial [\bm{X}_{p}]_{b,c}}{\partial [\bm{X}_{4}]_{b,c,d}}
\end{align*}
\vspace{-0.6em}
\begin{equation}
\begin{aligned}
    &=\frac{1}{BD}\sum_{n}[-w_{cn_b}+\frac{\exp( Y_{bn})}{\sum_{q}\exp(Y_{bq})}w_{cn}].
\end{aligned}
\end{equation}


We conduct experiments and record on $C_{\bm{Z}}$, $L_{\bm{\theta}}$ and $H_{\bm{\theta}}$ for a single minimal cell by fixing $\bm{\alpha}$ and and uniformly sampling networks in DSNAS, updating $\bm{\theta}$ for 150 epochs (Fig. \ref{fig:compare_dsnas_cost_appx}). We also fix $\bm{\alpha}$ and update $\bm{\theta}$ for 50 epochs in DARTS (Fig. \ref{fig:compare_darts_training_cost_appx}). The results validate this proof. Their dynamical trends also validate Eq. \ref{eq:cost_trend} and Remark \ref{thm:growth_trend}. 

\begin{figure}[h]
    \centering
    \includegraphics[width=2.8in]{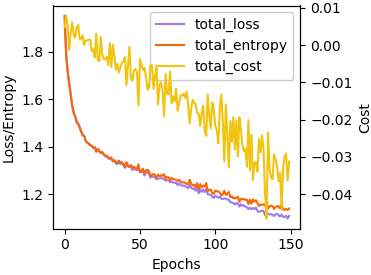}
    \caption{$C_{\bm{Z}}$, $L_{\bm{\theta}}$, $H_{\bm{\theta}}$ in DSNAS. X-axis is the epoch number, y-axis on the left is for $L_{\bm{\theta}}$ and $H_{\bm{\theta}}$ and the right is for $C_{\bm{Z}}$. Note that the unit for $C_{\bm{Z}}$ is magnified to show its trend.}
    \label{fig:compare_dsnas_cost_appx}
\end{figure}

\begin{figure}[h]
    \centering
    \includegraphics[width=2.8in]{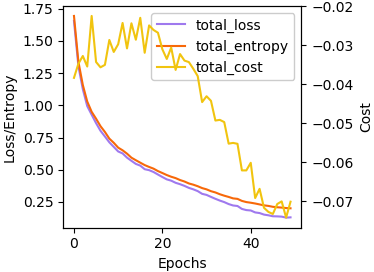}
    \caption{$C_{\bm{Z}}$, $L_{\bm{\theta}}$, $H_{\bm{\theta}}$ in DARTS in training set. X-axis is the epoch number, y-axis on the left is for $L_{\bm{\theta}}$ and $H_{\bm{\theta}}$ and the right is for $C_{\bm{Z}}$. Note that the unit for $C_{\bm{Z}}$ is magnified to show its trend.}
    \label{fig:compare_darts_training_cost_appx}
\end{figure}

\begin{figure}[h!]
    \centering
    \includegraphics[width=2.8in]{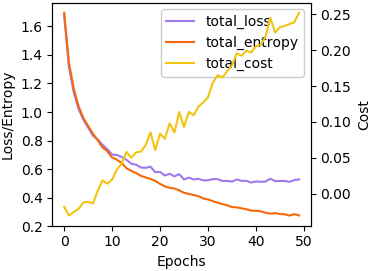}
    \caption{$C_{\bm{Z}}$, $L_{\bm{\theta}}$, $H_{\bm{\theta}}$ in DARTS in search set. X-axis is the epoch number, y-axis on the left is for $L_{\bm{\theta}}$ and $H_{\bm{\theta}}$ and the right is for $C_{\bm{Z}}$. Note that the unit for $C_{\bm{Z}}$ is magnified to show its trend.}
    \label{fig:compare_darts_search_cost_appx}
\end{figure}

Fig. \ref{fig:compare_darts_search_cost_appx} shows the result on the search set of DARTS. Note that the search set if a set of data used to train $\bm{\alpha}$, held-out from the training of $\bm{\theta}$. Curves in this figure illustrate our analysis in Remark \ref{thm:bi-level}. The loss decelerates its decreasing compared with the loss in training set (Fig. \ref{fig:compare_darts_training_cost_appx}), while the entropy keeps decreasing. 

\textbf{Positive at low accuracy}:
Exploiting normalization and weight initialization, we have:
\begin{equation*}
\label{eq:cost_sum_init_apx}
\begin{aligned}
\mathbb{E}_{\bm{\theta}_0}[\textcolor{red}{C_{\bm{Z}}}>0],
\end{aligned}
\end{equation*}
Note that in our derivation, we follow the commonly used parameter initialization method. With $[\bm{X}_{p}]_{b,c}\sim \mathcal{N}(0,\cdot)$, $w_{c,n}\sim \mathcal{N}(0,\cdot)$ at the initialization point, we have $y_{1},y_{2} \sim \mathcal{N}(0,\cdot)$ and $\mathbb{E}_{y_{1},y_{2} \sim \mathcal{N}(0,\cdot)}[y_{1}\exp{(y_{1}+y_{2}})]>0$, thus we can prove $\mathbb{E}_{\bm{\theta}_0}[C_{\bm{Z}}>0]$.

\textbf{Negative at high accuracy}: With operation parameters updated towards convergence, the probability of $b$-th image being classified to the correct label $n_b$ increases towards 1. 
Since $Y_{bn_b} = \max{\{Y_{bn}\}}$, we have
\begin{equation*}
\label{eq:cost_sum_update_apx}
\begin{aligned}
    \textcolor{red}{C_{\bm{Z}}} \propto & \textcolor{red}{\sum_{n}[-Y_{bn_b}+\frac{\exp( Y_{bn})}{\sum_{q}\exp(Y_{bq})}Y_{bn}]} \\
    \leq &\sum_{n}[-Y_{bn_b}+\frac{\exp( Y_{bn})}{\sum_{q}\exp(Y_{bq})}Y_{bn_b}] = 0.
\end{aligned}
\end{equation*}

Fig. \ref{fig:compare_dsnas_correct_wrong_cost_appx}, Fig. \ref{fig:compare_darts_training_correct_wrong_cost_appx} and Fig. \ref{fig:compare_darts_search_correct_wrong_cost_appx} show the comparison of $L_{\bm{\theta}}$ and $H_{\bm{\theta}}$ in the training set and the search set, which validate this proof. Note that the increase of the negative cost for correct classification in later epochs is mainly because the probability of $b$-th image being classified to the correct label $n_b$ increases towards 1.
\begin{figure}[h]
    \centering
    \includegraphics[width=3in]{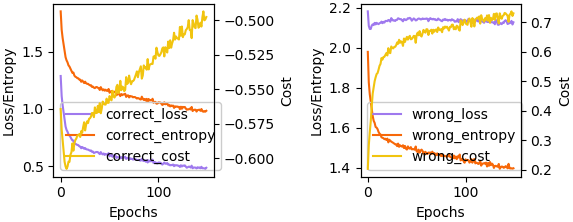}
    \caption{$C_{\bm{Z}}$, $L_{\bm{\theta}}$, $H_{\bm{\theta}}$ in DSNAS for (a) correct classification and (b) wrong classification.}
    \label{fig:compare_dsnas_correct_wrong_cost_appx}
\end{figure}

\begin{figure}[h!]
    \centering
    \includegraphics[width=3in]{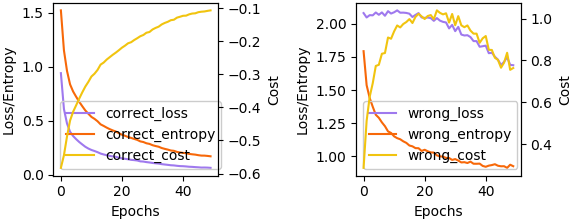}
    \caption{$C_{\bm{Z}}$, $L_{\bm{\theta}}$, $H_{\bm{\theta}}$ in DARTS for (a) correct classification and (b) wrong classification in training set}
    \label{fig:compare_darts_training_correct_wrong_cost_appx}
\end{figure}

\begin{figure}[h]
    \centering
    \includegraphics[width=3in]{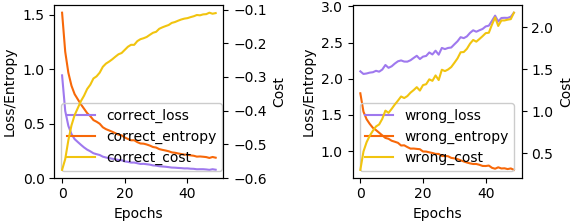}
    \caption{$C_{\bm{Z}}$, $L_{\bm{\theta}}$, $H_{\bm{\theta}}$ in DARTS for (a) correct classification and (b) wrong classification in search set}
    \label{fig:compare_darts_search_correct_wrong_cost_appx}
\end{figure}
\end{proof}

\section{Details for empirical study in Sec. \ref{sec:cost_of_inter_edge}}
\label{app:comp_details}

\begin{figure}[h]
    \centering
    \includegraphics[width=3in]{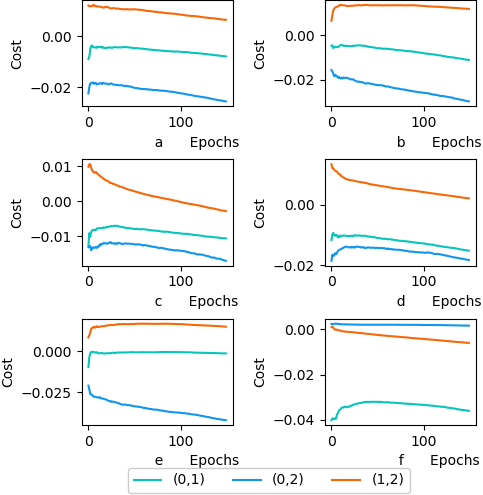}
    \caption{Averaged cost on each edge by (a) sampling none/skip/pool/conv on $edge (0,1)$, $edge(0,2)$, $edge(1,2)$, (b) sampling none/skip/pool/conv on $edge(0,1)$, $edge(0,2)$ and sampling none/skip/pool on $edge(1,2)$, (c) sampling none/skip/pool/conv on $edge(0,1)$, $edge(0,2)$ and sampling none/conv on $edge(1,2)$, (d) sampling none/skip/pool/conv on $edge(0,2)$, $edge(1,2)$ and sampling none/conv on $edge(0,1)$, (e) sampling none/skip/pool/conv on $edge(0,2)$, $edge(1,2)$ and sampling none/pool/skip on $edge (0,1)$, (f) sampling none/skip/pool/conv on $edge(1,2)$ and fixing operation on $edge(0,1)$, $edge(0,2)$. For almost all cases, $edge(1,2)$ has the greatest cost.}
    \label{fig:fig10a_cost_appx}
\end{figure}

\begin{figure}[h!]
    \centering
    \includegraphics[width=3in]{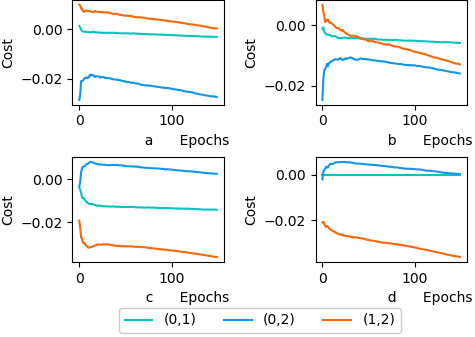}
    \caption{Averaged cost on each edge by deleting $edge (1,3)$ and (a) sampling none/skip/pool/conv on $edge (0,1)$, $edge (0,2)$, $edge (1,2)$, (b) sampling none/skip/pool/conv on $edge (0,2)$, $edge (1,2)$ and sampling skip/pool/conv on $edge (0,1)$, (c) sampling none/skip/pool/conv on $edge  (0,2)$, $edge (1,2)$ and fixing operation on $edge (0,1)$, (d) sampling none/skip/pool/conv on $edge (0,2)$ and sampling none/pool/conv on $edge (1,2)$ and fixing operation on $edge (0,1)$. Only in (c) and (d) $edge (1,2)$ has the lowest cost. The cost on $edge (0,1)$ being close to zero here also validates our proof of Thm. \ref{thm:bn0} in Appx. \ref{app:proof_thm2}.}
    \label{fig:fig10b_cost_appx}
\end{figure}

In Sec. \ref{sec:cost_of_inter_edge}, we introduced our empirical study to illustrate the distinctive role of intermediate edges. Here we provide more details to help readers understand our design of ablation. Fig. \ref{fig:fig10a_cost_appx} shows the cost of edges in simplified cell (Fig. \ref{fig:simplified_modified_cell}(a)), averaged over operations. The variation here is mainly different operation combination on edges, including fixing to one operation. Ones can see that $edge(1,2)$ always has the largest cost. Fig. \ref{fig:fig10b_cost_appx} shows experiments on modified cell (Fig. \ref{fig:simplified_modified_cell}(b)), where $edge (1,3)$ is deleted. When we still sample operation on $edge(0,1)$, the cost on $edge(1,2)$ is still larger than $edge(0,1)$ and $edge(0,2)$, as shown in Fig. \ref{fig:fig10b_cost_appx}(a)\&(b). This is only altered by also fixing the operation on $edge(0,1)$. In sum, the existence of $edge (1,3)$ and the sampling on $edge (0,1)$ are two major factors causing the discrimination towards the intermediate edge. 

Interestingly, even though in DARTS \textit{None} is excluded and a \textit{post-hoc} derivation scheme that select edges with top-k $\alpha$ is designed, intermediate edges are barely chosen. This can be explained by the observations here. Since the cost on intermediate edges are positive most of the time, their $\alpha$ must be smaller than other edges. 

\section{Further discussion}
\label{app:other_norm}
\subsection{Other Normalization Methods}
Ones may wonder if the inductive bias discussed in the work could be avoid by shifting to other normalization units, \textit{i.e.} instance normalization \citep{ulyanov2016instance}, layer normalization \citep{ba2016layer}. However, they seem no option than BN. The cost of instance normalization (affine-enabled) and layer normalization (affine-enabled) is the same as that of batch normalization. As an example, we show that layer normalization still satisfies Thm. \ref{thm:bn0}. The proof of instance normalization is similar.

Proof: The layer normalization process is shown as below:
\begin{equation}
\begin{aligned}
    &\mu_{b} = \frac{1}{CD}\sum_{c,d}\bm{X}_{b,c,d}, 
    &\sigma_{b}^{2} = \frac{1}{CD}\sum_{c,d}\left(\bm{X}_{b,c,d}-\mu_b\right)^2,\\
    &[\hat{\bm{X}}^{2}_{3}]_{b,c,d} = \frac{\bm{X}_{b,c,d}-\mu_b}{\sqrt{\sigma_{b}^{2}}}, 
    &[\bm{X}^{2}_{3}]_{b,c,d}=\gamma_b*[\hat{\bm{X}}^{2}_{3}]_{b,c,d}+\beta_b.
\end{aligned}
\end{equation}

Similarly, we have:
\begin{equation}
\label{eq:forward_layernorm}
\begin{aligned}
    &\frac{\partial L_{\bm{\theta}}}{\partial \bm{X}_{3}^{2}} \frac{\partial \bm{X}_{3}^{2}}{\partial \bm{X}}\bm{X}=\sum_{b,c,d}\frac{\partial L_{\bm{\theta}}}{\partial [\bm{X}^{2}_{3}]_{b,c,d}}\frac{\gamma_b\bm{X}_{b,c,d}}{\sqrt{\sigma_b^2}}-\frac{1}{CD}\sum_{b,c,d}(\sum_{c,d}\frac{\partial L_{\bm{\theta}}}{\partial [\bm{X}^{2}_{3}]_{b,c,d}})\frac{\gamma_b \bm{X}_{b,c,d}}{\sqrt{\sigma_b^2}}\\
    &-\sum_{b,c,d}\frac{\gamma_b}{CD}(\sum_{c,d}\frac{\partial L_{\bm{\theta}}}{\partial [\bm{X}^{2}_{3}]_{b,c,d}}[\hat{\bm{X}}^{2}_{3}]_{b,c,d})[\hat{\bm{X}}^{2}_{3}]_{b,c,d}\frac{\bm{X}_{b,c,d}}{\sqrt{\sigma_b^2}}\\
    &=\sum_{b,c,d}\frac{\partial L_{\bm{\theta}}}{\partial [\bm{X}^{2}_{3}]_{b,c,d}}\frac{\gamma_b(\bm{X}_{b,c,d}-\mu_b)}{\sqrt{\sigma_b^2}}-\mathunderline{orange}{\frac{1}{CD}\sum_{b,c,d}(\sum_{c,d}\frac{\partial L_{\bm{\theta}}}{\partial [\bm{X}^{2}_{3}]_{b,c,d}})\frac{\gamma_b(\bm{X}_{b,c,d}-\mu_b)}{\sqrt{\sigma_b^2}}}\\
    &-\sum_{b,c,d}\frac{\gamma_b}{CD}(\sum_{c,d}\frac{\partial L_{\bm{\theta}}}{\partial [\bm{X}^{2}_{3}]_{b,c,d}}[\hat{\bm{X}}^{2}_{3}]_{b,c,d})[\hat{\bm{X}}^{2}_{3}]_{b,c,d}\frac{\bm{X}_{b,c,d}}{\sqrt{\sigma_b^2}}\\
    &=\sum_{b,c,d}\frac{\partial L_{\bm{\theta}}}{\partial [\bm{X}^{2}_{3}]_{b,c,d}}\gamma_b[\hat{\bm{X}}^{2}_{3}]_{b,c,d}-\sum_{b,c,d}\frac{\gamma_b}{CD}(\sum_{c,d}\frac{\partial L_{\bm{\theta}}}{\partial [\bm{X}^{2}_{3}]_{b,c,d}}[\hat{\bm{X}}^{2}_{3}]_{b,c,d})[\hat{\bm{X}}^{2}_{3}]_{b,c,d}\frac{\bm{X}_{b,c,d}}{\sqrt{\sigma_b^2}}\\
    &=\sum_{b}\gamma_b(\sum_{c,d}\frac{\partial L_{\bm{\theta}}}{\partial [\bm{X}^{2}_{3}]_{b,c,d}}[\hat{\bm{X}}^{2}_{3}]_{b,c,d})(1-\sum_{c,d}\frac{[\hat{\bm{X}}^{2}_{3}]_{b,c,d}^2}{CD})\\
    &=0,
\end{aligned}
\end{equation}
where the \textcolor{orange}{\textbf{orange-underlined}} term is strictly 0 due to the \textbf{standard Gaussian mean property} $\sum_{c,d}\frac{[\hat{\bm{X}}^{2}_{3}]_{b,c,d}}{CD}=0$, the last equality also holds due to \textbf{standard Gaussian variance property} $\sum_{c,d}\frac{[\hat{\bm{X}}^{2}_{3}]^2_{b,c,d}}{CD} = 1$.

\begin{figure}[h]
    \centering
    \includegraphics[width=3in]{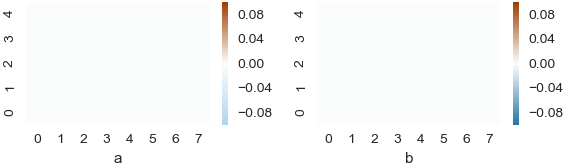}
    \caption{Cost mean statistics for each operation (x-axis) on each edge (y-axis) by using affine-disabled instance normalization.}
    \label{fig:insnorm_result_appx}
\end{figure}

Interestingly, the cost mean statistic collected on different edges by using affine-disabled instance normalization are exactly zero, and they will still be zero after training. Fig. \ref{fig:insnorm_result_appx} shows the cost mean statistic.

We can derive the cost function $C(Z_{i,j}^{s})$ by using affine-disabled instance normalization. Based on the path (4-3-0) (or (4-3-1), (4-3-2)), we write the cost function explicitly on $edge 2$ ($edge 3$, $edge 4$). The cost function on $edge 2$ ($edge 3$, $edge 4$) is derived to be zero for all operations. 
\begin{equation}
\label{eq:cost_edge34_ln_appx}
\begin{aligned}
    C(Z_{i,j}^{s})&=\sum_{b,c,d}\frac{\partial L_{\bm{\theta}}}{\partial [\bm{X}_{4}^{3}]_{b,c,d}} [\bm{X}_{3}^{0}]_{b,c,d}\\
    &=\frac{1}{D}\sum_{b,c}\frac{\partial L_{\bm{\theta}}}{\partial [\bm{X}_{p}]_{b,c}} \sum_{d}[\bm{X}_{3}^{0}]_{b,c,d}\\
    &=0
\end{aligned}
\end{equation}
where $\sum_{d}[\bm{X}_{4}^{3}]_{b,c,d}=0$ (instance normalization property).

The cost on $edge 0$ ($edge 1$) can be derived similarly and follows the same conclusion as $edge 2$ ($edge 3$, $edge 4$). 
 
\subsection{Stacking order in operations}
Another consideration could be to change the order of Conv, BN and ReLU units in operations. In our proposed framework, we can show that both theoretically and empirically, same phenomenon would occur. 

\begin{figure}[h!]
    \centering
    \includegraphics[width=3in]{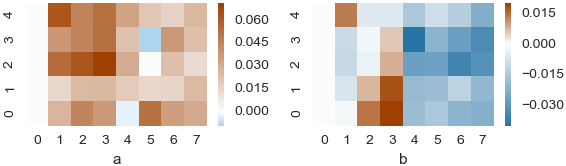}
    \caption{Cost mean statistics for each operation (x-axis) on each edge (y-axis) (a) at initialization and (b) near convergence of $\theta$ by using \textit{Conv-ReLU-BN} order. Operation 0: none, Operation 1: skip connect, Operation 2: max\_pool\_3x3, Operation 3: avg\_pool\_3x3, Operation 4: sep\_conv\_3x3, Operation 5: dil\_conv\_3x3, Operation 6: dil\_conv\_5x5, Operation 7: sep\_conv\_5x5.}
    \label{fig:Conv_ReLU_bn_appx}
\end{figure}

Fig. \ref{fig:Conv_ReLU_bn_appx} shows the cost mean statistic on different edges by using \textit{Conv-ReLU-BN} or \textit{Pooling-BN} order in the minimal cell structure. 

\begin{figure}[h!]
    \centering
    \includegraphics[width=3in]{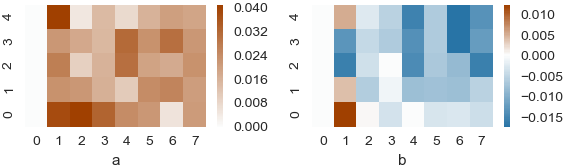}
    \caption{Cost mean statistics for each operation (x-axis) on each edge (y-axis) (a) at initialization and (b) near convergence of $\theta$ by using \textit{Conv-BN-ReLU} order. Operation 0: none, Operation 1: skip connect, Operation 2: max\_pool\_3x3, Operation 3: avg\_pool\_3x3, Operation 4: sep\_conv\_3x3, Operation 5: dil\_conv\_3x3, Operation 6: dil\_conv\_5x5, Operation 7: sep\_conv\_5x5.}
    \label{fig:Conv_bn_ReLU_appx}
\end{figure}

\vspace{-4mm}
Fig. \ref{fig:Conv_bn_ReLU_appx} shows the cost mean statistic on different edges by using \textit{Conv-BN-ReLU} or \textit{Pooling-BN} order in the minimal cell structure. 

The proof in Appx.\ref{app:proof_thm2} and Appx.\ref{app:proof_thm3} are also valid after the change of order, since the order of \textit{ReLU} operation does not make a difference.

\end{document}